%% file: article.tex
\documentclass[10pt,twocolumn,letterpaper]{article}

\usepackage{iccv}
\usepackage{times}
\usepackage{epsfig}
\usepackage{graphicx}
\usepackage{amsmath}
\usepackage{amssymb}
\usepackage{bm}
\usepackage{multirow}
\usepackage{booktabs}
\usepackage{enumitem}
\usepackage{lipsum}
\usepackage{ntheorem}
\usepackage{float}
\usepackage{dsfont}
\usepackage{siunitx}

\setlist[itemize]{leftmargin=12pt}

\usepackage[format=plain,labelformat=simple,labelsep=period,font=small,compatibility=false,skip=3pt]{caption}
\usepackage[font=footnotesize,skip=3pt,subrefformat=parens]{subcaption}

\usepackage[pagebackref=true,breaklinks=true,letterpaper=true,colorlinks,bookmarks=false]{hyperref}

\theoremstyle{break}
\newtheorem{property}{Property}
\newtheorem{appproperty}{Property}
\newtheorem{proposition}{Proposition}
\newtheorem{appproposition}{Proposition}
\theoremstyle{nonumberplain}
\newtheorem{proof}{Proof}

\usepackage[capitalize]{cleveref}
\crefname{section}{Sec.}{Secs.}
\Crefname{section}{Section}{Sections}
\Crefname{table}{Table}{Tables}
\crefname{table}{Tab.}{Tabs.}
\crefname{property}{Property}{Properties}
\crefname{appproperty}{Property}{Properties}
\crefname{appendix}{App.}{Apps.}

\iccvfinalcopy 

\def\iccvPaperID{10852} 

\ificcvfinal\fi

\begin{document}

\title{Rethinking Intersection Over Union for Small Object Detection in \\Few-Shot Regime}

\author{Pierre Le Jeune\\
L2TI, Université Sorbonne Paris Nord\\
COSE\\
{\tt\small pierre.le-jeune@cose.fr}
\and
Anissa Mokraoui\\
L2TI, Université Sorbonne Paris Nord\\
First line of institution2 address\\
{\tt\small anissa.mokraoui@univ-paris13.fr}
}

\maketitle
\ificcvfinal\thispagestyle{empty}\fi

\begin{abstract}
   In Few-Shot Object Detection (FSOD), detecting small objects is extremely
   difficult. The limited supervision cripples the localization capabilities of
   the models and a few pixels shift can dramatically reduce the Intersection
   over Union (IoU) between the ground truth and predicted boxes for small
   objects. To this end, we propose Scale-adaptive Intersection over Union
   (SIoU), a novel box similarity measure. SIoU changes with the objects' size,
   it is more lenient with small object shifts. We conducted a user study and
   SIoU better aligns than IoU with human judgment. Employing SIoU as an
   evaluation criterion helps to build more user-oriented models. SIoU can also
   be used as a loss function to prioritize small objects during training,
   outperforming existing loss functions. SIoU improves small object detection
   in the non-few-shot regime, but this setting is unrealistic in the industry
   as annotated detection datasets are often too expensive to acquire. Hence,
   our experiments mainly focus on the few-shot regime to demonstrate the
   superiority and versatility of SIoU loss. SIoU improves significantly FSOD
   performance on small objects in both natural (Pascal VOC and COCO datasets)
   and aerial images (DOTA and DIOR). In aerial imagery, small objects are
   critical and SIoU loss achieves new state-of-the-art FSOD on DOTA and DIOR. 
\end{abstract}

\input{Parts/01_introduction.tex}

\input{Parts/02_existing_work.tex}

\input{Parts/03_proposed_criterion.tex}

\input{Parts/04_criterion_analysis.tex}

\input{Parts/05_results.tex}

\input{Parts/06_conclusion.tex}

{\small
\bibliographystyle{ieee_fullname}
\bibliography{egbib}
}
\newpage
\appendix
\input{Parts/appendices.tex}

\end{document}

%% file: Parts/01_introduction.tex
\section{Introduction}
\label{sec:intro}

Object detection is a fundamental task in industry and has applications in many
domains such as medical imaging, agriculture, and autonomous driving. However,
it is often impracticable or too expensive to build sufficiently large annotated
datasets to train detection models. It is therefore crucial to improve
data-efficient approaches and particularly Few-Shot Object Detection (FSOD)
methods. However, the limited number of examples provides poor supervision and
prevents the model to learn accurate localization, which is especially
problematic for small objects. Besides, the difficulty of detecting small
objects was already reported in many object detectors
\cite{ren2015faster,redmon2016you, liu2016ssd,
tian2019fcos,duan2019centernet,law2018cornernet,carion2020end}. Numerous
attempts partially solved this issue by proposing various improvements such as
pyramidal features \cite{liu2016ssd, lin2017feature, zhang2018single } or
multiscale training \cite{singh2018sniper, singh2021scale}. However, this
difficulty greatly intensifies in the few-shot regime as shown by
\cite{lejeune2022improving}. One of the reasons for the poor FSOD performance on
small objects is the extensive use of Intersection over Union (IoU). Most
detection (and so FSOD) pipelines employ IoU as a regression loss
\cite{yu2016unitbox, tian2019fcos}; for example selection
\cite{ren2015faster,redmon2016you, liu2016ssd}; or an evaluation criterion, but
IoU is not an optimal choice when dealing with small objects.

\begin{figure}
    \centering
    \begin{subfigure}[t]{0.23\textwidth}
        \vskip 0pt
        \centering
        \includegraphics[trim=0 5 0 5, clip,width=\textwidth]{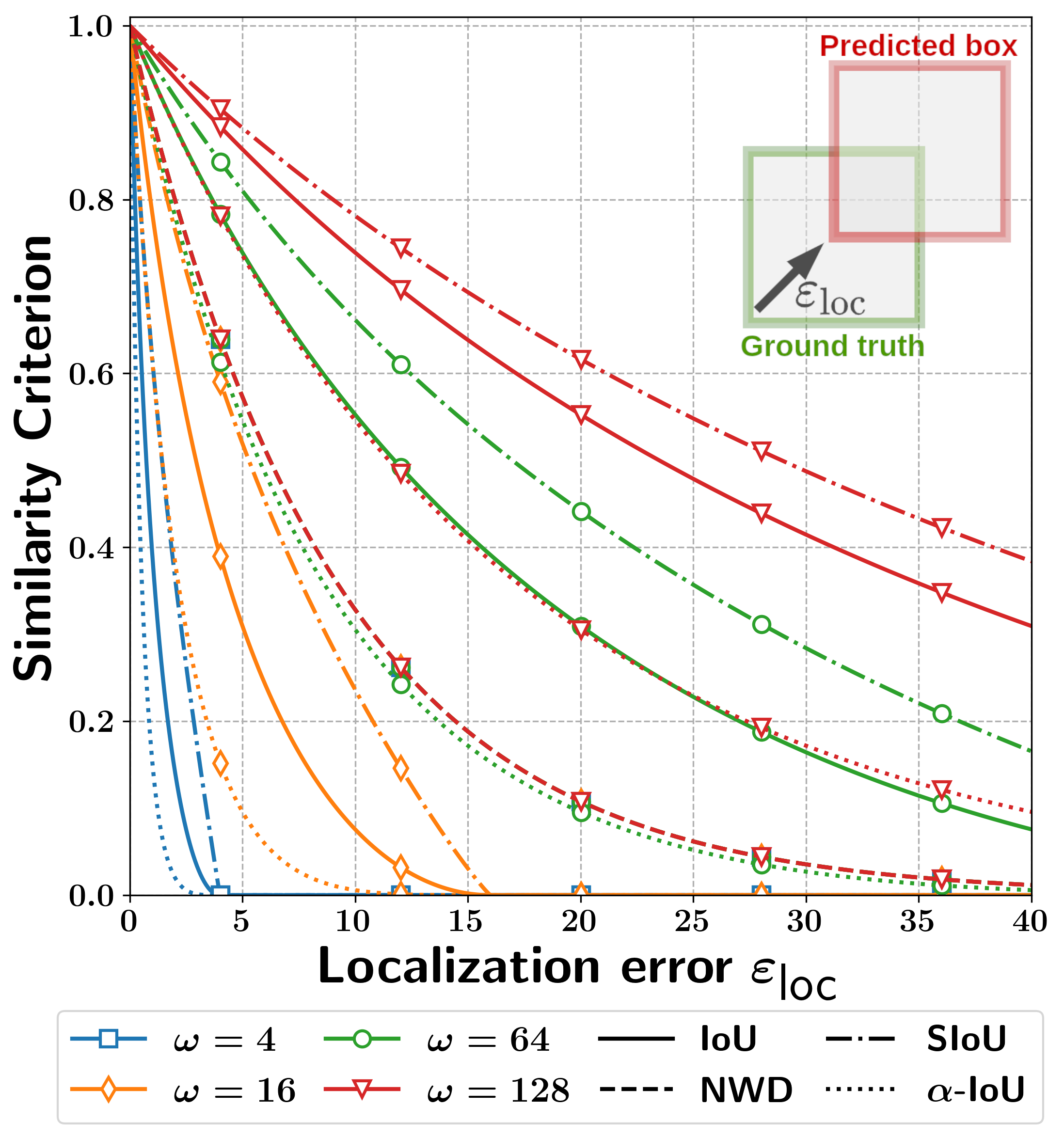}
        \label{fig:pixel_shift}
    \end{subfigure}
    \begin{subfigure}[t]{0.24\textwidth}
        \vskip 0pt
        \centering
        \includegraphics[trim=0 0 0 7, clip,width=0.99\textwidth]{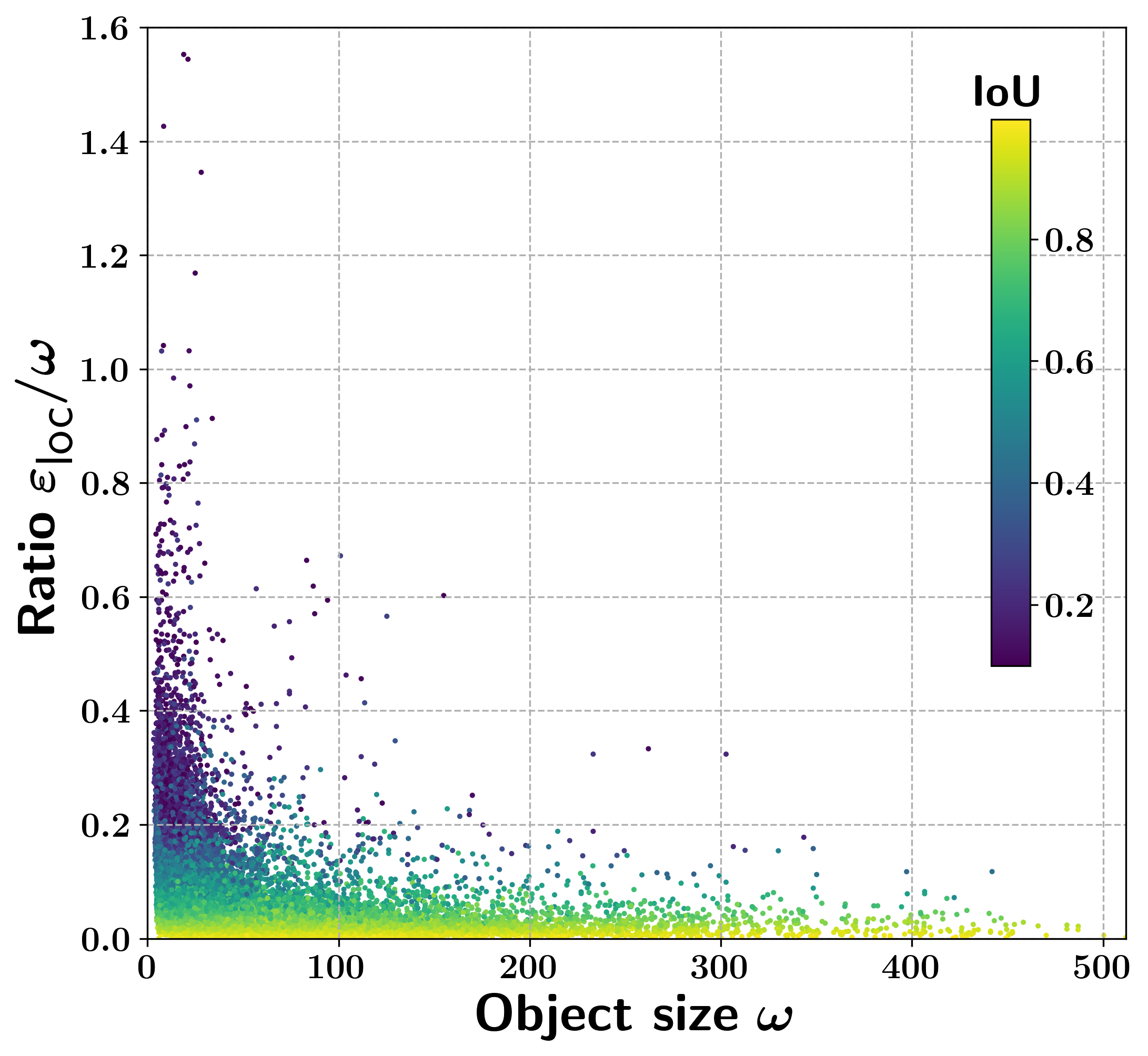}
        \vskip 6pt
        \label{fig:model_inaccuracy}
    \end{subfigure}
    \vspace{-4mm}
    \caption{\textbf{(Left)} Evolution of IoU, NWD \cite{wang2021nwd}, the
    proposed SIoU and $\alpha$-IoU \cite{he2021alpha} when a box is shifted from
    the ground truth box by $\varepsilon_{\text{loc}}$ pixels for various box
    sizes $\omega \in \{4, 16, 64, 128\}$. \textbf{(Right)} Ratio between pixel
    localization error $\varepsilon_{\text{loc}}$ and object size $\omega$ for a
    trained detection model on DOTA dataset. Each point represents the
    localization error of one object in DOTA test set.}
    \label{fig:loc_error_evo}
\end{figure}

IoU has a remarkable property: scale invariance. It means that scaling all
coordinates of two bounding boxes by the same amount will not change their IoU.
At first glance, this seems a desirable property, all objects will be treated
identically no matter their size. In practice, it has a fundamental drawback:
small boxes are prone to large IoU changes from only small position or size
modifications. To clarify, let us consider a simple example. Two
square boxes of width $\omega$ are shifted diagonally by
$\varepsilon_{\text{loc}}$ pixels. In this setup, a 1-pixel shift leads to a
larger decrease in IoU when boxes are small. This comes from the scale
invariance property, IoU stays constant as the ratio
$\frac{\varepsilon_{\text{loc}}}{\omega}$ remains fixed. 
However, this ratio is not constant for trained detection models, it increases
as objects get smaller (see \cref{fig:loc_error_evo} right), leading to lower
IoU values for smaller objects. Hence, small objects are much more likely to
fall under the IoU thresholds which decide if a box is a true or false
detection, even though being satisfactory from a human perspective (see the user
study in \cref{sec:user_study}). \cref{sec:shift_analysis,sec:inaccuracy}
explore the resilience of various criteria to localization inaccuracies and
confirm that IoU is not an optimal box similarity measure. 

Only a handful of works question the adequation of IoU for object detection.
Among those, \cite{rezatofighi2019generalized} proposed a generalization of IoU
when boxes do not overlap, \cite{wang2021nwd} introduced a novel loss function
to target small objects and \cite{strafforello2022humans} showed that human
perception and IoU are not fully aligned. This lack of interest in new criterion
design is explained by the great detection performance in the regular setting
(\ie natural images with sufficient annotations). In the few-shot regime, and
when targets are small, the flaws of IoU become critical. Therefore,         
we revisit IoU to improve FSOD methods and focus on aerial images which mostly
contain small objects. We propose Scale-adaptive Intersection over
Union (SIoU), a novel criterion that can replace IoU for training and evaluating
detection models. To demonstrate the superiority of the proposed SIoU,
\cref{sec:criterion_analysis} compares it with various existing criteria. This
section analyzes the criteria's distributions when exposed to randomly shifted
boxes. To our knowledge, this is the first attempt to study
empirically and theoretically the distributions of these criteria. The
conclusions of this analysis are then compared with human perception through a
user study which shows that SIoU aligns better with human appraisal than IoU
(see \cref{sec:user_study}). The comparison of the criteria also highlights that
SIoU as a loss function can guide training towards small objects better than
other criteria and in a more controlled fashion. SIoU loss can be tuned to
improve the detection of small objects just as it can be tuned to align with
human perception. Finally, these analyses are confirmed by extensive experiments
on both aerial images (DOTA \cite{xia2018dota} and DIOR \cite{li2020dior}
datasets) and natural images (Pascal VOC \cite{everingham2010pascal} and COCO
\cite{lin2014microsoft} datasets).\\

\noindent
The main contributions of this paper are as follows:

\begin{itemize}[noitemsep]
    \item[-] A novel scale-adaptive criterion called SIoU that can be tuned to detect
    objects of various sizes.
    \item[-] An empirical and theoretical analysis of existing
    criteria that help to understand the required properties for designing
    regression loss functions and evaluation criteria.
    \item[-] A user study that demonstrates the misalignment between IoU and human
    perception for the detection task.
    \item[-] Extensive experiments to support the superiority of
    SIoU for detecting small objects in the few-shot regime. 
\end{itemize}


%% file: Parts/02_existing_work.tex
\section{Related Works}
\label{sec:rel_work}

\subsection{Intersection over Union and its Variants}
To begin, let us review the definition of existing criteria for set similarity.
First, the IoU is defined as the
intersection area of two sets divided by the area of their union: 

\begin{equation}
    \text{IoU}(A, B) = \frac{|A \cap B|}{|A \cup B|},
\end{equation}

\noindent
where $A$ and $B$ are two sets. When $A$ and $B$ are rectangular boxes, IoU can
be computed easily with simple operations on box coordinates. This explains
why IoU is such a widespread criterion for object detection. It is used as a
loss function ($\mathcal{L}_{\text{reg}} = 1 - \text{IoU}$) by several well
established detection frameworks (\eg \cite{yu2016unitbox, tian2019fcos}). IoU
is also involved in the process of example selection during training of most
detection methods, \ie all the ones inspired either by Faster R-CNN
\cite{ren2015faster} or YOLO \cite{redmon2016you}. In these frameworks,
regression loss is computed from the coordinates of proposed boxes and ground
truth. Not all pairs of proposals and ground truth are kept for the computation.
Only proposals with a sufficient IoU with a ground truth box are selected.
Finally, IoU is also used at the heart of the evaluation process. A proposed box
is considered a positive detection if it meets two conditions: 1) an IoU greater
than a given threshold with a ground truth box, and 2) the same label as this
ground truth. 

Several attempts were made to improve IoU but existing works mostly focus on the
regression loss part, disregarding the other IoU uses in the detection
task. First, \cite{rezatofighi2019generalized} proposed a generalized version of
IoU which yields negative values when boxes do not overlap: 
\begin{equation}
    \text{GIoU}(A,B) = \text{IoU}(A,B) - \frac{|C \backslash (A \cup B)|}{|C|},
\end{equation}

\noindent
where $C$ is the convex hull around $A$ and $B$. This criterion is employed as a
loss function by several detection frameworks \cite{tian2019fcos,
bochkovskiy2020yolov4, zhang2020bridging}. It is sometimes also combined with
other regression loss as in \cite{li2020generalized, carion2020end}, which both
combine it with an L1 regression on box coordinates. Combining IoU loss with
other regression terms was also proposed by \cite{zheng2020distance}. They introduce
two losses Distance-IoU (DIoU) and Complete-IoU which respectively add an L2
regression term and an aspect ratio penalty to the IoU loss. Recently,
$\alpha$-IoU \cite{he2021alpha} extends DIoU \cite{zheng2020distance} by
proposing a family of losses following the same structure as DIoU with the IoU
term raised to the power $\alpha$. Alternatively, Bounded IoU
\cite{tychsen2018improving} computes an IoU upper bound between a proposal and a
ground truth. 

All previous IoU improvements were made to tackle the regression part of the
models. However, IoU is involved in other parts of the framework including
example selection, Non-Maximal Suppression, and evaluation. A recent user
study \cite{strafforello2022humans} indicates that IoU does not completely align
with human perception. Humans have strong positional and size preferences based on
conceptual information contained in the boxes. It suggests that IoU is not
an optimal choice either for example selection or for evaluation as it
will lead to detections that do not satisfy human users.  

\subsection{Object Detection}
Object Detection is a problem that has been studied for decades. It witnessed
rapid progress with the rise of deep learning methods
\cite{ren2015faster,redmon2016you, liu2016ssd}. Recent methods achieve very
satisfactory results when provided with sufficient data. However, there remain
some challenges to mastering object detection. Most object detectors still
struggle with small objects, and when data is scarce.

\noindent
\textbf{Small Object Detection} is a challenging task. There has been plenty of
attempts to improve it based on pyramidal features \cite{liu2016ssd,
lin2017feature, zhang2018single }, multiscale training \cite{singh2018sniper,
singh2021scale}, data-augmentation \cite{kisantal2019augmentation} or
super-resolution \cite{rabbi2020small, Bai2018SODMTGANSO, Shermeyer2019TheEO,
Courtrai2020SmallOD,Ferdous2019SuperRD}. But only a few works tackle this
problem by changing the loss function. Normalized Wasserstein Distance
\cite{wang2021nwd} (NWD) proposes an alternative to IoU loss specifically
designed for detecting small objects. It consists in computing the Wasserstein
distance between two Gaussian distributions fitted on the two compared bounding
boxes. Moreover, NWD is also used as an example selection criterion.

\noindent
\textbf{Few-Shot Object Detection (FSOD)} is the task of detecting objects only
provided with a handful of examples per class. Many approaches were proposed in the
literature to address this problem: metric learning
\cite{karlinsky2019repmet,yang2020restoring, sun2021fsce, jeune2021experience},
simple fine-tuning \cite{chen2018lstd, wang2020frustratingly, wolf2021double,
xu2022simpl} and attention-based methods \cite{kang2019few, xiao2020fsod,
han2021meta, zhang2021meta, chen2021should}. The similarity between all these
methods is that they learn generic knowledge from a set of \textit{base classes}
with plenty of annotations and adapt to \textit{novel classes} from the few
available examples. Recently, it has been shown \cite{lejeune2022improving} that
FSOD is even more sensitive to small objects than regular object detection.
Extracting information from small objects is hard and produces spurious features
that do not condition well the detection. Some solutions are proposed to
overcome this issue with augmentation and careful example cropping
\cite{lejeune2022improving} or with dedicated attention mechanisms
\cite{lejeune2022aaf}. Nevertheless, this is not enough to solve the issue of
small objects in FSOD.


%% file: Parts/03_proposed_criterion.tex
\section{Novel Scale-Adaptive Intersection over Union}
\label{sec:siou}

Before introducing the proposed criterion, let us define two bounding boxes
$b_1=[x_1, y_1, w_1, h_1]^T$ and $b_2 = [x_2, y_2, w_2,
h_2]^T$ (the prediction box and ground truth respectively), where $x_i$ and
$y_i$ are the center coordinates of the box $b_i$, while $w_i$ and $h_i$ denote its
width and height respectively. In the following section, the adjectives small,
medium, and large will be used extensively. They have a precise meaning for
object detection, defined in COCO dataset \cite{lin2014microsoft}. The box $b_i$
is \textit{small} if $\sqrt{w_ih_i} \leq 32$ pixels, \textit{medium} if $32 <
\sqrt{w_ih_i} \leq 96$, and \textit{large} if $\sqrt{w_ih_i} > 96$.

IoU is scale-invariant, hence if $\text{IoU}(b_1,b_2) =
u$, scaling all coordinates of both boxes by the same factor $k$ will produce
the same IoU: $\text{IoU}(b_1, b_2) =
\text{IoU}(kb_1, kb_2) = u$. However, detection models are
not scale-invariant, they do not localize equally well small and large objects.
\cref{fig:loc_error_evo} (right) clearly shows that the ratio between the
localization error ($\varepsilon_{\text{loc}} = \| b_1 - b_2\|_1 $) and the
object size ($\omega=\sqrt{w_2h_2}$) increases as the objects become
smaller. This figure is made with a model trained on DOTA with all annotations.
Each point represents the ratio $\frac{\varepsilon_{\text{loc}}}{\omega}$ for
one object in the test set. Hence, because of the scale-invariance property, IoU
scores are lower for small objects. A way to alleviate this issue is by relaxing
the invariance property of the IoU so it favors more small objects without
penalizing large ones. To this end, we propose a novel criterion called
Scale-adaptive Intersection over Union (SIoU): 
\begin{equation}
    \begin{aligned}
        \text{SIoU}(b_1, b_2) &= \text{IoU}(b_1, b_2) ^ p  \label{eq:p}\\
        \text{with} \quad\quad p &= 1 - \gamma \exp\left({-\frac{\sqrt{w_1h_1 + w_2h_2}}{\sqrt{2}\kappa}}\right),
    \end{aligned}
\end{equation}

\noindent
$p$ is a function of the object sizes, thus, the scores are rescaled according
to the objects' size. $\gamma \in ]-\infty, 1]$ and $\kappa > 0$ are two
parameters that control how the rescaling occurs (hence, $p \geq 0$). $\gamma$
governs the scaling for small objects while $\kappa$ controls how fast the
behavior of regular IoU is recovered for large objects.
\cref{fig:influence_gamma_kappa} (left) in \cref{app:gamma_kappa} shows the
evolution of $p$ with object size for various $\gamma$ and $\kappa$.

This new criterion follows the same structure as $\alpha$-IoU
\cite{he2021alpha}, but differs greatly as it sets
different powers for different object sizes. SIoU provides a solution for small
object detection while $\alpha$-IoU only aims to improve general detection.
However, SIoU inherits a few properties from $\alpha$-IoU. 

\begin{property}[SIoU Relaxation]
    \label{property:relaxation}
    Let $b_1$ and $b_2$ be two bounding boxes and introduce $\tau = \frac{w_1h_1 +
    w_2h_2}{2}$ their average area. SIoU preserves the behavior of IoU in
    certain cases such as:
   \begin{itemize}
    \setlength\itemsep{0em}
        \item[-]$\textup{IoU}(b_1, b_2) = 0 \Rightarrow \textup{SIoU}(b_1, b_2) =\textup{IoU}(b_1, b_2) = 0$
        \item[-]$\textup{IoU}(b_1, b_2) = 1 \Rightarrow\textup{SIoU}(b_1, b_2) =\textup{IoU}(b_1, b_2) = 1$
        \item[-]$\lim\limits_{\tau \to +\infty} \textup{SIoU}(b_1, b_2) =\textup{IoU}(b_1,b_2)$
        \item[-]$\lim\limits_{\kappa\to 0} \textup{SIoU}(b_1, b_2) =\textup{IoU}(b_1,b_2)$
   \end{itemize}
\end{property}

\cref{property:relaxation} shows that SIoU is sound: it equals IoU when boxes have no
intersection and when they perfectly overlap. Therefore, the associated loss
function (see \cref{property:l_g_reweight}) will take maximal values for boxes that
do not overlap and minimum ones for identical boxes. In addition, SIoU
behaves similarly to IoU when dealing with large objects (\ie when $\tau \to \infty$).
When boxes are large, the power $p$ that rescales the IoU is close to 1. Hence,
this change of criterion only impacts small objects. However, when discussing
the properties of SIoU, the limit between small/medium/large objects is relative
to the choice of $\kappa$. If $\kappa \gg \sqrt{wh}$, even large objects will be
rescaled. On the contrary, when $\kappa \xrightarrow {}0$, all objects are
treated as large and are not rescaled. In practice, $\kappa$ and $\gamma$ are
chosen empirically, but \cref{sec:criterion_analysis} provides useful insights
for the choice of these parameters.  

\begin{property}[Loss and gradients reweighting]
    \label{property:l_g_reweight}
    Let $\mathcal{L}_{\textup{IoU}}(b_1, b_2) = 1 - \textup{IoU}(b_1, b_2)$ and
    $\mathcal{L}_{\textup{SIoU}}(b_1, b_2) = 1 - \textup{SIoU}(b_1, b_2)$ be the
    loss functions associated respectively with IoU and SIoU. Let denote the
    ratio between SIoU and IoU losses by 
    $\mathcal{W}_{\mathcal{L}}(b_1, b_2) =
    \frac{\mathcal{L}_{\textup{SIoU}}(b_1, b_2)}{\mathcal{L}_{\textup{IoU}}(b_1,
    b_2)}$. 
    Similarly, 
    $\mathcal{W}_{\mathcal{\nabla}}(b_1, b_2) =
    \frac{|\nabla\mathcal{L}_{\textup{SIoU}}(b_1, b_2)|}{|\nabla\mathcal{L}_{\textup{IoU}}(b_1,
    b_2)|}$ 
    denotes the ratio of gradients generated from SIoU and IoU losses: 
    \begin{align}
        \mathcal{W}_{\mathcal{L}}(b_1, b_2) &= \frac{1- \textup{IoU}(b_1, b_2)^p}{1-\textup{IoU}(b_1, b_2)}, \\
        \mathcal{W}_{\mathcal{\nabla}}(b_1, b_2) &= p\textup{IoU}(b_1, b_2)^{p-1},
    \end{align}

    \noindent
    $\mathcal{W}_{\mathcal{L}}$ and $\mathcal{W}_{\mathcal{\nabla}}$ are
    increasing (resp. decreasing) functions of IoU when $p\geq 1$ (resp. $p <
    1$) which is satisfied when $\gamma \leq 0$ (resp. $\gamma > 0$). As the IoU
    goes to 1, $\mathcal{W}_{\mathcal{L}}$ and $\mathcal{W}_{\mathcal{\nabla}}$
    approaches $p$: 
    \begin{align}
        \lim\limits_{\textup{IoU}(b_1, b_2) \to 1}\mathcal{W}_{\mathcal{L}}(b_1, b_2) &=  p, \\
        \lim\limits_{\textup{IoU}(b_1, b_2) \to 1}\mathcal{W}_{\mathcal{\nabla}}(b_1, b_2) &=  p.
    \end{align}

\end{property}

We employ the same tools as in \cite{he2021alpha} to analyze how SIoU affects
the losses and associated gradients. We show in property 2 that their results
hold for a non-constant power $p$ as well. From this, it can be observed that
when IoU is close to 1, losses and gradients are both rescaled by $p$. Hence,
the gradients coming from objects of different sizes will be rescaled differently.
The setting of $\gamma$ and $\kappa$ allows focusing the training on specific
object sizes. Experimental results are provided in \cref{sec:results} to
support these findings. Proofs for properties 1 and 2 are available in
\cref{app:properties}    

However, \textit{order preservingness} is not satisfied by using power value
changing with the size of the objects. This property ensures that the order
given by the IoU is preserved with the novel criterion, \eg $\text{IoU}(b_1,b_2) <
\text{IoU}(b_1,b_3) \Rightarrow \alpha\text{-IoU}(b_1,b_2) <
\alpha\text{-IoU}(b_1,b_3)$. $\alpha$-IoU preserves the order of IoU, but SIoU
does not. We show in \cref{app:properties} that even though this property is not
always satisfied, a large proportion of boxes meet the conditions for the order
to hold. \\


\noindent
\textbf{Extensions and generalization}\\
Finally, SIoU can very well be extended as IoU was with GIoU or DIoU. We provide
here an extension following GIoU as it appears especially well-designed for
small object detection. When detecting small targets, it is easier for a model
to completely miss the object, producing an IoU of 0 no matter how far the
predicted box is. On the contrary, GIoU yields negative values for
non-intersecting boxes. This produces more relevant guidance during the early phase
of training when the model outputs poorly located boxes. Therefore, we extend
SIoU by raising GIoU to the same power $p$ as in \cref{eq:p}:%
\begin{equation}
    \text{GSIoU}(b_1, b_2) = \begin{cases}\text{g}(b_1, b_2) ^ p  &\text{if } \text{g}(b_1, b_2) \geq 0 \\
                                            -|\text{g}(b_1, b_2)| ^ p  &\text{if } \text{g}(b_1, b_2) < 0 \end{cases}, 
\end{equation}

\noindent
where $g(b_1, b_2)=\text{GIoU}(b_1,b_2)$.

%% file: Parts/04_criterion_analysis.tex
\section{Scale-Adaptive Criteria Analysis}
\label{sec:criterion_analysis}

\begin{figure*}
    \centering
    \includegraphics[width=\textwidth]{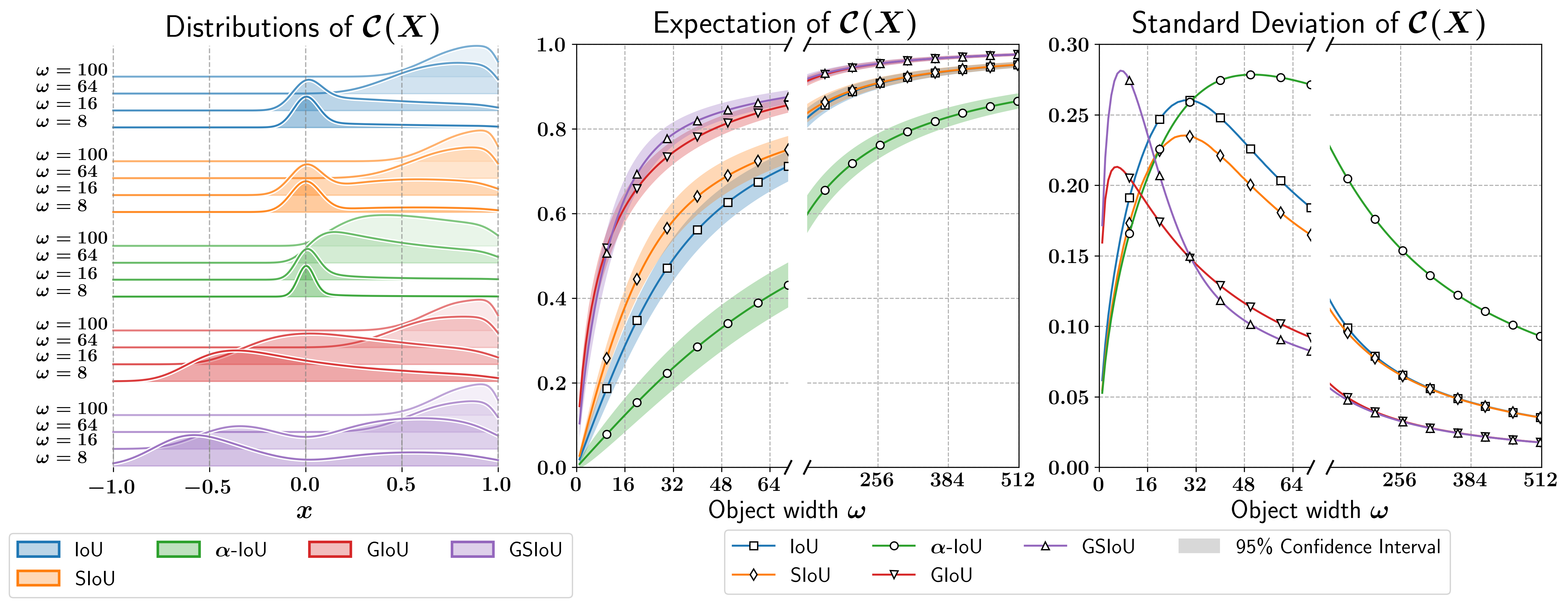}
    \caption{Analysis of the distribution of IoU, SIoU, GIoU, GSIoU and
            $\alpha$-IoU when computed on inaccurately positioned boxes. This is
            done by observing the probability distribution functions for various
            $\omega$ values \textbf{(left)}, the expectation \textbf{(middle)}
            and standard deviation \textbf{(right)} for all criteria. For SIoU
            and GSIoU, we fixed $\gamma=0.5$ and $\kappa=64$, for $\alpha$-IoU,
            $\alpha=3$ (as recommended in the original paper
            \cite{he2021alpha}). The inaccuracy of the detector is set to
            $\sigma=16$. Note that the empirical pdfs were smoothed using a
            Kernel Density Estimator method. This affects particularly IoU, SIoU
            , and $\alpha$-IoU for the actual pdf is defined only on $[0,1]$. For
            the sake of visualization, GIoU and GSIoU were rescaled between 0
            and 1 for the expectation and standard deviation plots. }
    \label{fig:criteria_analysis}
\end{figure*}

This section analyzes both empirically and theoretically the behaviors of IoU,
GIoU \cite{rezatofighi2019generalized}, $\alpha$-IoU \cite{he2021alpha}, NWD
\cite{wang2021nwd}, SIoU and GSIoU. We investigate the desirable properties of
such criteria for model training and performance evaluation.

\subsection{Response Analysis to Box Shifting}
\label{sec:shift_analysis}
As mentioned in \cref{sec:siou}, IoU drops dramatically when the localization
error increases for small objects. Shifting a box a few pixels off the ground
truth can result in a large decrease in IoU, without diminishing the quality of
the detection from a human perspective. This is depicted in
\cref{fig:loc_error_evo} (left), plain lines represent the evolution of IoU for
various object sizes. These curves are generated by diagonally shifting a box
away from the ground truth. Boxes are squares, but similar curves would be
observed otherwise. In this plot, boxes have the same size, therefore, when
there is no shift in between ($\varepsilon_{\text{loc}}=0$), IoU equals 1.
However, if the sizes of the boxes differ by a ratio $r$, IoU would peak at
$1/r^2$. Other line types represent other criteria. SIoU decreases slower than
IoU when $\varepsilon_{\text{loc}}$ increases, this is especially true when
boxes are small. This holds because $\gamma=0.5$, if it was negative, SIoU would
adopt the opposite behavior. In addition, the gap between IoU and
SIoU is even larger when objects are small. Only NWD shares this property, but
it only appears when boxes have different sizes (all lines coincide for NWD).
Hence, SIoU is the only criterion that allows controlling its decreasing rate,
\ie how much SIoU is lost for a 1-pixel shift. As GIoU and GSIoU values range in
$[-1,1]$, they were not included in \cref{fig:loc_error_evo}, but the same
analysis holds for them as well (see \cref{app:inaccuracy}).

\subsection{Resilience Analysis to Detector Inaccuracy}
\label{sec:inaccuracy}
Knowing how a criterion responds to shifts and size variations is important to
understand what makes a sensible box similarity measure. Pushing beyond the shift
analysis, we study empirically and theoretically the criteria's distributions
when exposed to detector inaccuracies, \ie randomly shifted boxes. This setting
mimics the inaccuracy of the model either during training or at test time.

\subsubsection{Empirical Protocol}
\label{sec:empirical_analysis}
To simplify, let us suppose that all boxes are squares of the same size $\omega$
and can be shifted only horizontally. Similar results are observed by
relaxing these constraints, see \cref{app:inaccuracy}. A box is then entirely
defined by its position $x$ and its width $\omega$. If a detector is not
perfect, it will produce bounding boxes slightly shifted horizontally from the
ground truth. To model the detector's inaccuracy, we suppose that the box position
is randomly sampled from a centered Gaussian distribution:
$X \sim \mathcal{N}(0, \sigma^2)$ where $\sigma$ controls how inaccurate the model
is. We are interested in the distribution of $\mathcal{C} \in
\{\text{IoU}, \text{GIoU}, \text{SIoU}, \text{GSIoU}, \alpha\text{-IoU},
\text{NWD}\}$ and how it changes with $\omega$. To this end, let $Z =
\mathcal{C}(X)$. More precisely, we are interested in the probability density function
(pdf) of $Z$ and its two first moments (which exist because $\mathcal{C}$
is continuous and bounded). 

\cref{fig:criteria_analysis} gathers the results of this analysis. It shows the
pdf of each criterion for various box sizes (left) along with the evolution of
the expectation and standard deviation of $Z$ against $\omega$ (middle and
right). From this, it can be noted that the size of the boxes has a large
influence on the distributions of all criteria. The expected values of all
criteria are monotonically increasing with object size. In particular, small
objects have lower expected IoU values than larger ones. This is consistent with
the initial assessment from \cref{fig:loc_error_evo} (right) and it validates
the choice of $\sigma$ constant for this study (although \cref{app:inaccuracy}
discusses this assumption).

When building detection models, we hope to detect equally well objects of all
sizes, this means having a constant expected IoU, no matter the objects' size.
This would require the localization error to be an affine function of $\omega$.
Of course, the localization error of the detector is likely to depend on
$\omega$. However, it cannot be an affine function, otherwise, small objects
would be perfectly detected, which is not observed (see
\cref{fig:loc_error_evo}, right). As SIoU has larger expected values than IoU
for small objects, it can compensate for their larger localization errors. The
setting of $\gamma$ and $\kappa$ allows controlling how much small objects are
favored (see \cref{fig:influence_gamma_kappa} in \cref{app:gamma_kappa}). NWD
is not included in these plots as its expected value and variance are constant
when dealing with same-size boxes.\\

\noindent
\textbf{Influence Analysis on the Performance Evaluation}\\
If the expected value of a criterion is too small, it is likely that the boxes
will be considered negative detections during evaluation and therefore reduce
the performance. Therefore, having a criterion with larger expected values for
small objects would better reflect the true performance of a detector. One might
think that it would be equivalent to scale-adaptive IoU thresholds during the
evaluation, but this is not completely true as the variances of the criteria also
differ.

Having an accurate criterion (\ie with low variance) is crucial for evaluation.
Let us take a detector that produces well-localized boxes on average, \ie on
average the criterion computed between the boxes and their corresponding ground
truths is above a certain threshold. As the detector is not perfect, it will
randomly produce boxes slightly better or slightly worse than the average. If
the criterion has a high variance, it will be more likely that poor boxes get
scores below the criterion threshold and therefore will be considered 
negative detections. This will reduce the performance of the detector even
though on average, it meets the localization requirements. In addition, a
criterion with a higher variance will be less reliable and would produce more
inconsistent evaluations of a model. The fact that the IoU variance is high for
small objects partly explains why detectors have much lower performance on these
objects. Hence, SIoU seems more adapted for evaluation. Of course,
using this criterion for evaluation will attribute higher scores for less
precise localization of small objects. However, this aligns better with human
perception as demonstrated in \cref{sec:user_study}. Employing SIoU as a metric
also allows tweaking that metric for the needs of a specific application. \\ 

\noindent
\textbf{Influence Analysis on Training}\\
All criteria discussed above are employed as regression losses in the
literature. The loss associated with each criterion $\mathcal{C}$ is
$\mathcal{L}_{\mathcal{C}}(b_1,b_2) = 1 - \mathcal{C}(b_1,b_2)$. Therefore, the
expected value of the criterion determines the expected value of the loss and
thus the magnitude of the gradients. Large values of the criterion give low
values of the loss. Now, as the expected values of the criteria change
with the objects size, the expected values of the losses also change. Small
objects generate greater loss values than larger ones on average. However, this is
not enough as performance on small objects is poor when training with IoU.
To achieve better detection, training must focus even more on small objects. One
way to do this is to set larger loss values for small objects. That way,
the equilibrium is shifted toward smaller objects and gradients will point to
regions where the loss of small objects is lower. As shown in
\cref{fig:criteria_analysis_4} (\cref{app:gamma_kappa}), with the right
parameters, SIoU can do that. It attributes lower values for small objects while
keeping similar values for large ones. The contrast between small and large
objects is accentuated and optimization naturally focuses on smaller
objects. SIoU's parameters control which object size gets
more emphasis. This is closely linked to \cref{property:l_g_reweight} which
states that employing SIoU (compared to IoU) reweights the loss and the gradient
by $p$. If $\gamma < 0$, $p$ decreases with the size of the objects and thus the
optimization focuses on small objects. This also explains why generalizations of
existing criteria (\ie with negative values for non-overlapping boxes) often
outperform their vanilla version. Taking IoU and GIoU as an example, the gap
between their expected values for small and large objects is greater with GIoU.
It nudges the optimization towards small objects.  

\subsubsection{Theoretical study of GIoU}
Criteria pdfs and first moments can also be derived theoretically. We provide
such results for GIoU in \cref{prop:distribution_giou}. 

\begin{proposition}[GIoU's distribution]
    \label{prop:distribution_giou}
    Let $b_1 = (0,y_1,w_1,h_1)$ be a bounding box horizontally centered and $b_2
    = (X,y_2,w_2,h_2)$ another bounding box randomly positioned, with $X\sim
    \mathcal{N}(0, \sigma^2)$ and $\sigma \in \mathbb{R}^*_+$. Let's suppose that
    the boxes are identical squares, shifted only horizontally (\ie $w_1 = w_2 = h_1 =
    h_2$ and $y_1 = y_2$). \\
    \noindent
    Let $Z = \mathcal{C}(X)$, where $\mathcal{C}$ is the generalized intersection over union. The
    probability density function of $Z$ is given by:
    \begin{align}
        d_Z(z) &= \frac{2\omega}{(1+z)^2\sqrt{2\pi} \sigma} \exp\left(-\frac{1}{2}\left[\frac{\omega(1-z)}{\sigma(1+z)}\right]^2\right).
    \end{align}

    \noindent
    The two first moments of $Z$ exist and are given by:
    \begin{align}
        \mathbb{E}[Z] &= \frac{2}{\pi^{3/2}} G^{2,3}_{3,2}\left(2a^2 \bigg\rvert \begin{matrix}0 & \frac{1}{2} & \frac{1}{2} \\ \frac{1}{2}& 0 \end{matrix}\right), \\
        \mathbb{E}[Z^2] &= 1- \frac{8a}{\sqrt{2\pi}} + \frac{16a^2}{\pi^{3/2}} G^{2,3}_{3,2}\left(2a^2 \bigg\rvert \begin{matrix}-1 & \frac{1}{2} & -\frac{1}{2} \\ \frac{1}{2}  & 0 \end{matrix}\right),
    \end{align}

    \noindent
    where $G$ is the Meijer G-function \cite{meijer2013} and $a=\frac{\sigma}{\omega}$. 
\end{proposition}
The proof of this
proposition and derivations for other criteria are available in
\cref{app:inaccuracy}. The theoretical expressions completely agree with
empirical results, which confirms the soundness of our simulations. 

\subsection{SIoU Alignment with Human Perception}
\label{sec:user_study}
As discussed in \cref{sec:empirical_analysis}, having an accurate criterion \ie
one with low variance, is crucial for evaluation. However, such a criterion
must also align with human perception. Most image processing models
are destined to assist human users. Therefore, to maximize the usefulness of such
models, the evaluation process should align as closely as possible with human
perception. To assess the agreement between the criteria and human perception,
we conducted a user study in which participants had to rate on a 1 to 5 scale
(\ie from \textit{very poor} to \textit{very good}) how a bounding box detects an
object. Specifically, an object is designated by a green ground truth box and a
red box is randomly sampled around the object (\ie with random IoU with the
ground truth). Then, the participants rate how well the red box localizes the
object within the green one. The study gathered 75 different participants and
more than 3000 individual answers. We present here the main conclusion of this
study. Detailed results and protocol are available in
\cref{app:user_study}. 

\begin{figure}
    \centering
    \includegraphics[width=\columnwidth]{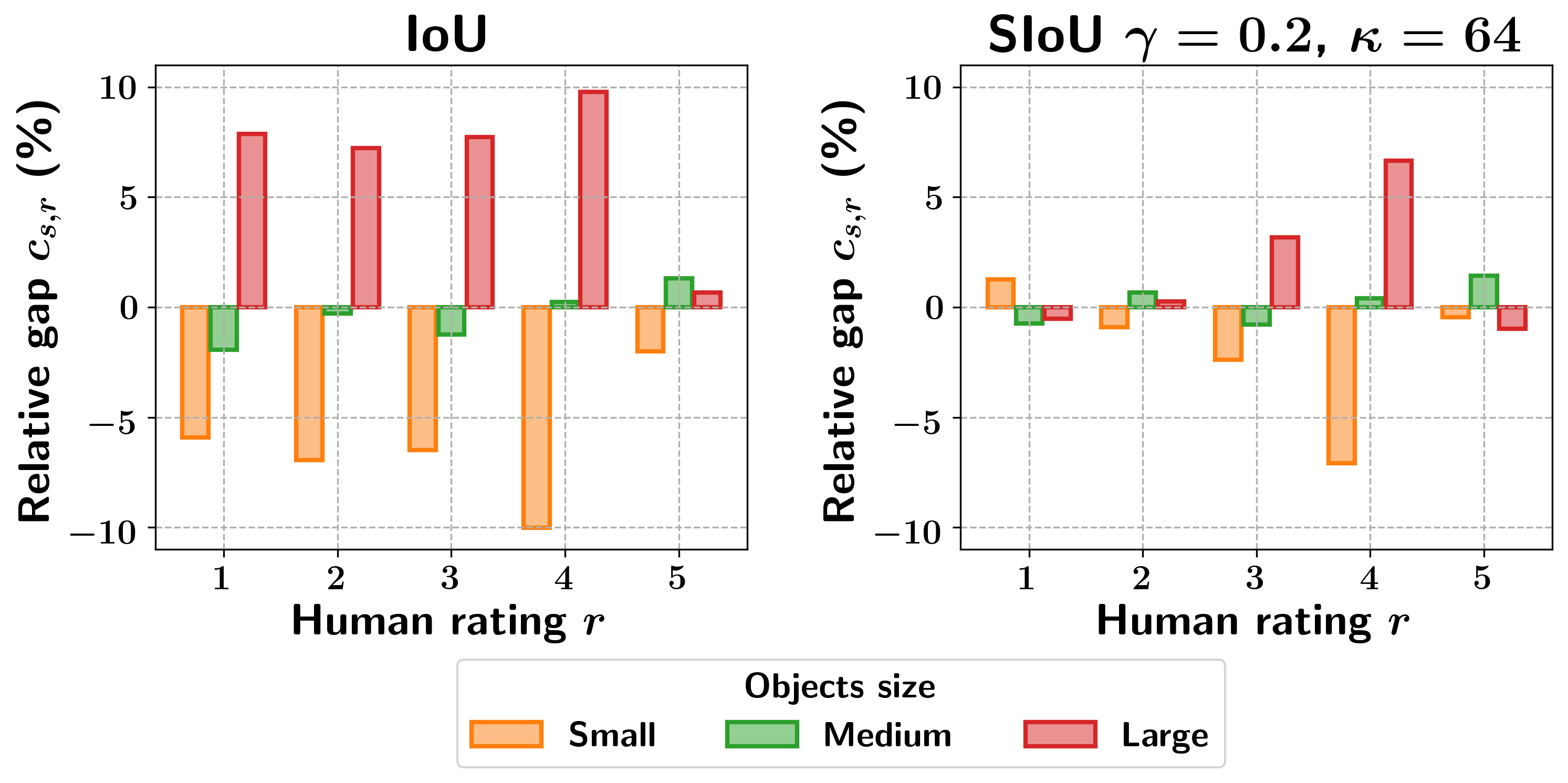}
    \caption{Average IoU (left) and SIoU (right) scores for different object
    sizes and human ratings $r \in \{1,2,3,4,5\}$. Values are reported as
    the relative gap with the average value per rating.}
    \label{fig:rating_vs_crit}
\end{figure}

Human perception does not fully align with IoU. People tend to be more lenient
than IoU towards small objects. Specifically, comparing a small and a large box
with the same IoU with respect to their own ground truth, people will rate the
small one better. This suggests that IoU is too strict for small objects in
comparison with human perception. From a human perspective, precise localization
seems less important for small objects. \cref{fig:rating_vs_crit} represents the
relative gap of IoU (left) and SIoU (right) values for each object size and
rating. The relative differences $c_{s,r}$ are computed against the average IoU (or SIoU)
value per rating: 
\begin{equation}
    c_{s,r} =
\frac{\mathcal{C}_{s,r} -
\sum\limits_s\mathcal{C}_{s,r}}{\sum\limits_s\mathcal{C}_{s,r}},
\end{equation}
\noindent
where $\mathcal{C}_{s,r}$ is the average criterion value ($\mathcal{C} \in
\{\text{IoU}, \text{SIoU}\}$) for objects of size $s$ and rating $r$. IoU values
for small objects (in orange) are lower than for large objects (in red) for all
rating $r$. For a human to give a rating $r$ to a box, it requires that a box
overlaps less with the ground truth (according to IoU) if the boxes are small.
SIoU compensates for this trend (see \cref{fig:rating_vs_crit} (right)): SIoU
differences between objects of different sizes but the same rating are smaller
than for IoU. This means that SIoU process objects independently of their size.
Similar charts with absolute IoU and SIoU values can be found in
\cref{app:user_study} (\cref{fig:bar_chart_user_study}), it also includes charts
for $\alpha$-IoU, NWD and other SIoU parameters. Here $\gamma = 0.2$ and
$\kappa=64$ for SIoU. Choosing a higher $\gamma$ value would reverse the trend
and produce a criterion even more lenient than humans for small objects. It will
also decrease further SIoU's variance. However, this setting has been chosen to
maximize the alignment with human perception. SIoU correlates better with human
rating compared with other criteria. As the rating is an ordered categorical
variable, we choose Kendall rank correlation to make the comparison. The
correlation between the human rating $r$ and each criterion can be found in
\cref{tab:correlation} (correlations between criteria are available in
\cref{tab:avg_rating} in \cref{app:user_study}). SIoU with $\gamma=0.2$ and
$\kappa=64$ aligns best with human perception and has a low variance. This
showcases the superiority of SIoU over existing criteria. It should be preferred
over IoU to assess the performance of models on all visual tasks that commonly
employ IoU within their evaluation process. It supports recent findings that
show misalignment between IoU and human preference
\cite{strafforello2022humans}.


\begin{table}[]
    \centering
    \resizebox{0.5\columnwidth}{!}{%
    \begin{tabular}{@{}ccccc@{}}
    \toprule
    \textbf{}                   & \textbf{IoU} & \textbf{SIoU} & $\boldsymbol{\alpha}$\textbf{-IoU} & \textbf{NWD} \\ \midrule
    $\boldsymbol{r}$          & 0.674        & \textbf{0.701}& 0.674               & 0.550        \\ \bottomrule
    \end{tabular}%
    }
    \caption{Kendall's $\tau$ correlation between various criteria and human
    rating $r$. For SIoU, $\gamma=0.2$ and $\kappa=64$, for $\alpha$-IoU,
    $\alpha=3$.}
    \vspace{-2mm}
    \label{tab:correlation}
\end{table}

%% file: Parts/05_results.tex
\section{Experimental Results}
\label{sec:results}

To support our analysis from \cref{sec:criterion_analysis}, we conduct various
experiments, mainly on aerial images with DOTA \cite{xia2018dota} and DIOR
\cite{li2020dior} datasets. To showcase the versatility of SIoU, we also
experiment with natural images on Pascal VOC \cite{everingham2010pascal} and
COCO \cite{lin2014microsoft}. Detecting small objects in the few-shot regime is
extremely challenging and could have much more applications than regular
detection pipelines. Therefore, most of our experiments focus on the few-shot
setting. However, we also report results in regular object detection to display
the potential of SIoU. For the few-shot experiments, we choose a recently
proposed FSOD method: Cross-Scale Query Support Alignment (XQSA)
\cite{lejeune2022aaf}. A comparison with other methods is available in
\cref{app:fs_comparison}. Since the FSOD training is relatively complex, we
defer the implementation details in \cref{app:implem_details}. In few-shot
literature, it is common to evaluate models separately on base and novel
classes, however novel classes performance is what matters the most as it
assesses the generalization capabilities of the models. Performance is computed
using mean Average Performance (mAP) with a 0.5 IoU threshold.

\noindent
\textbf{Comparison with Existing Criteria} \\
To begin, we compare the few-shot performance on DOTA with various loss
functions designed with the criteria discussed in \cref{sec:criterion_analysis}.
The result of these experiments is available in
\cref{tab:criterion_comparison}. The criteria
are divided into two groups, generalized (which includes NWD) and vanilla
criteria. As discussed in \cref{sec:empirical_analysis}, the generalized versions
of the criteria outperform their original counterparts and therefore should be
compared separately. Scale-adaptive criteria (SIoU and GSIoU) largely
outperforms other criteria on novel classes and especially on small objects. For
SIoU and GSIoU, we choose $\gamma=-3$ and $\kappa=16$ according to a series of
experiments conducted on DOTA to determine their optimal values (see
\cref{app:gamma_kappa}). It is important to point out the relatively good
performance of NWD despite not checking all the desirable properties highlighted
in \cref{sec:criterion_analysis}.

\begin{table}[]
    \centering
    \resizebox{\columnwidth}{!}{%
    \begin{tabular}{@{}cccccccccc@{}}
    \toprule[1.5pt]
    \multicolumn{1}{l}{\textbf{}} & \multicolumn{4}{c}{\textbf{Base classes}}          & & \multicolumn{4}{c}{\textbf{Novel Classes}}          \\ 
    \multicolumn{1}{l}{\textbf{Loss}} & \textbf{All} & \textbf{S} & \textbf{M} & \textbf{L}& & \textbf{All} & \textbf{S} & \textbf{M} & \textbf{L} \\ \midrule[1.5pt]
    \textbf{IoU}                  & 50.67        & \textbf{25.83}      & 57.49      & 68.24     & & 32.41        & 10.06      & 47.87      & 67.09      \\
    $\alpha$\textbf{-IoU}         & 46.72        & 13.24      & 55.21      & \textbf{69.94}     & & 33.95        & 12.58      & 46.58      & \textbf{74.50}      \\ 
    \textbf{SIoU}                 & \textbf{53.62}        & 24.07      & \textbf{61.91}      & 67.34     & & \textbf{39.05}        & \textbf{16.59}      & \textbf{54.42}      & \textbf{74.49}      \\ \midrule
    \textbf{NWD}                  & 50.79        & 19.19      & 58.90      & 67.90     & & 41.65        & 28.26      & 50.16      & 65.06      \\
    \textbf{GIoU}                 & 52.41        & \textbf{26.94}      & 61.17      & 63.00     & & 41.03        & 24.01      & \textbf{52.13}      & 69.78      \\
    \textbf{GSIoU}                & \textbf{52.91}        & 22.14      & \textbf{61.19}      & \textbf{66.02}     & & \textbf{45.88}        & \textbf{34.83}      & 51.26      & \textbf{70.78}      \\ \bottomrule[1.5pt]
    \end{tabular}%
    }
    \caption{Few-shot performance comparison between several criteria: IoU,
    $\alpha$-IoU, SIoU, NWD, GIoU, and GSIoU trained on DOTA. mAP is reported
    with a 0.5 IoU threshold for small (S), medium (M), large (L), and all objects.}
    \label{tab:criterion_comparison}
    \vspace{-2mm}
    \end{table}

\noindent
\textbf{FSOD on Aerial and Natural Images} \\
As the previous set of experiments was only carried out on DOTA, we showcase the
versatility of GSIoU on three other datasets: DIOR, Pascal VOC and COCO. As it is clear
that generalized criteria outperform other methods, the comparison here is only
done between GIoU and GSIoU. For DOTA and DIOR, current state-of-the-art is
achieved by XQSA \cite{lejeune2022aaf}, which employs GIoU as regression loss.
Therefore, we replace it with GSIoU and achieve significantly better performance
on these two datasets (see \cref{tab:dataset_comparison}). 

For Pascal VOC and COCO, similar gains are observed, but it requires a different
tuning of SIoU. $\gamma=-3$ and $\kappa=16$ produce mitigated results these
datasets, and $\gamma=-1$ and $\kappa=64$ is a more sensible choice. This was
predictable as the objects in Pascal VOC and COCO are substantially larger than
in DOTA and DIOR. This can also explain the slightly smaller gains on DIOR
compared to DOTA. Finding optimal values of $\gamma$ and $\kappa$ could yield
slightly better performance on DIOR. The tuning of SIoU is quite
straightforward, as lower values of $\gamma$ and $\kappa$ skew the training
towards smaller objects. The right balance depends on the proportion of small,
medium and large objects in the datasets. With natural images which contain
fewer small objects, the training balance does not need to be shifted as much as
for aerial images. In addition to these results, we also conducted several
experiments with various FSOD methods to demonstrate the plug-and-play nature of
GSIoU. These results are available in \cref{app:experimental_results} and show
consistent improvements when replacing GIoU with GSIoU.  

\begin{table}[]
    \centering
    \resizebox{\columnwidth}{!}{%
    \begin{tabular}{@{}ccccccccccc@{}}
    \toprule[1.5pt]
    \textbf{}                        & \textbf{} & \multicolumn{4}{c}{\textbf{Base classes}}          & & \multicolumn{4}{c}{\textbf{Novel Classes}} \\ 
    \textbf{}                        & \textbf{XQSA} & \textbf{All} & \textbf{S} & \textbf{M} & \textbf{L}& & \textbf{All}  & \textbf{S}  & M     & L    \\\midrule[1.5pt]
    \multirow{2}{*}{\textbf{DOTA}}   &  w/ GIoU      & 52.41        & \textbf{26.94}      & 61.17      & 63.00     & & 41.03          & 24.01         & \textbf{52.13}  & 69.78 \\
                                     &  w/ GSIoU     & \textbf{52.91}        & 22.14      & \textbf{61.19}      & \textbf{66.02}     & & \textbf{45.88}          & \textbf{34.83}         & 51.26  & \textbf{70.78} \\ \midrule
    \multirow{2}{*}{\textbf{DIOR}}   &  w/ GIoU      & 58.90        & 10.38      & 40.76      & 80.44     & & 47.93          & 9.85          & 47.61  & 68.40 \\
                                     &  w/ GSIoU     & \textbf{60.29}        & \textbf{11.28}      & \textbf{43.24}      & \textbf{81.63}     & & \textbf{52.85}          & \textbf{13.78}         & \textbf{53.73}  & \textbf{71.22} \\ \midrule
    \multirow{2}{*}{\textbf{Pascal}} &  w/ GIoU      & 51.09        & \textbf{13.93}      & \textbf{40.26}      & 62.01     & & 48.42          & 18.44         & 36.06  & 59.99 \\
                                     &  w/ GSIoU     & \textbf{54.47}        & 13.88      & 40.13      & \textbf{66.82}     & & \textbf{55.16}          & \textbf{22.94}         & \textbf{36.24}  & \textbf{67.40} \\ \midrule
    \multirow{2}{*}{\textbf{COCO}}&  w/ GIoU      & 19.15        & \textbf{8.72}     & 22.50      & 30.59                        & & 26.25          & 11.96         & 23.95  & 38.60 \\
                                     &  w/ GSIoU     & \textbf{19.57}        & 8.41      & \textbf{23.02}      & \textbf{31.07}     & & \textbf{27.11}          & \textbf{12.81}         & \textbf{26.02}  & \textbf{39.20} \\ \bottomrule[1.5pt]
    \end{tabular}%
    }
    \caption{Few-shot performance on three datasets: DOTA, DIOR, Pascal VOC and COCO.
    GIoU and GSIoU losses are compared. mAP is reported with a 0.5 IoU threshold
    and for all object sizes.}
    \vspace{-2mm}
    \label{tab:dataset_comparison}
    \end{table}

\noindent
\textbf{Regular Object Detection on DOTA and DIOR}\\
Of course, GSIoU is not only beneficial for FSOD, but it also improves the
performance of regular object detection methods. \cref{tab:regular_od} compares
the performance of FCOS \cite{tian2019fcos} trained on DOTA and DIOR with GIoU
and GSIoU. The same pattern is visible, we get better performance with GSIoU.
However, the gain for small objects is not as large as for FSOD. Nevertheless,
it suggests that other tasks relying on IoU could also benefit from GSIoU.

\begin{table}[]
    \centering
    \resizebox{0.9\columnwidth}{!}{%
    \begin{tabular}{@{}cccccccccc@{}}
    \toprule[1.5pt]
    \textbf{}      & \multicolumn{4}{c}{\textbf{DOTA}}                  & & \multicolumn{4}{c}{\textbf{DIOR}}          \\ 
    \textbf{FCOS}      & \textbf{All} & \textbf{S} & \textbf{M} & \textbf{L}& & \textbf{All} & \textbf{S} & \textbf{M} & L \\ \midrule
    \textbf{w/ GIoU}  & 34.9         & 17.4       & 36.6       & 43.3  & & 48.1        &10.1       &40.3       &63.2   \\
    \textbf{w/ GSIoU} & \textbf{36.8}         & \textbf{17.5}       & \textbf{40.4}       & \textbf{45.2}  & & \textbf{49.2}        & \textbf{11.0}      & \textbf{41.2}     &\textbf{66.1}   \\ \bottomrule[1.5pt]
    \end{tabular}%
    }
    \caption{Regular Object Detection performance on DOTA and DIOR datasets with
    GIoU and GSIoU ($\gamma=-3$ and $\kappa=16$) losses. mAP is computed with
    several IoU thresholds (0.5 to 0.95) as it is commonly done in regular
    detection.}
    \label{tab:regular_od}
    \end{table}

\noindent
\textbf{Discussions and Limitations}\\
As mentioned in \cref{sec:empirical_analysis} SIoU is a better choice for
performance analysis. However, as IoU is almost the only choice in literature
for evaluation, we must use it as well for a fair comparison with existing
works. Nonetheless, we provide results from previous tables using SIoU as the
evaluation criterion in \cref{app:siou_eval}. They agree with the IoU evaluation
and strengthen the conclusions of our experiments. While these results are
promising, we must emphasize a few limitations of SIoU. First, SIoU requires a
slight tuning to get the best performance, but that tuning is quite
straightforward and mostly depends on the size distribution in the target
images. SIoU allows being more lenient with small objects for evaluation
($\gamma \geq 0$), and stricter for training ($\gamma \leq 0$) to prioritize the
detection of small targets. Although they are not always part of the detection
pipeline, it would be relevant to investigate the replacement of IoU by SIoU for
example selection and Non-Maximal Suppression (in our case, the example
selection of FCOS does not rely on IoU). Finally, even though SIoU aligns better
than IoU with human perception, it does not match completely with it. IoU and
SIoU do not account for object content whereas humans heavily do, as highlighted
by \cite{strafforello2022humans}.

%% file: Parts/06_conclusion.tex
\section{Conclusion}
\label{sec:conclusion}
SIoU is a more suitable alternative than IoU both for object detection model evaluation
and training, especially in the few-shot regime. It better aligns with human
perception and our theoretical analysis confirms sounder properties for
evaluation. To our knowledge, this is the first statistical analysis of bounding
boxes similarity measure and hopefully, this will lead to more reflections on
object detection criteria. As a loss function, SIoU can incline the training
toward small objects and therefore greatly improve FSOD performance. Its
flexibility allows to easily focus the detection on specific target sizes and
adapt to various tasks. Extensive experiments on aerial and natural images
demonstrate the superiority of SIoU on small objects, without performance loss on
medium and large objects. On aerial images, which contain a lot of small
objects, SIoU even achieves state-of-the-art FSOD performance.

%% file: Parts/appendices.tex
\section{Influence of $\gamma$ and $\kappa$ on SIoU and GSIoU}
\label{app:gamma_kappa}
In this appendix, we analyze the influence of the parameters $\gamma$ and
$\kappa$ on the SIoU and GSIoU.

\begin{figure*}
   \centering
   \begin{subfigure}[t]{\textwidth}
       \centering
       \includegraphics[width=0.93\textwidth]{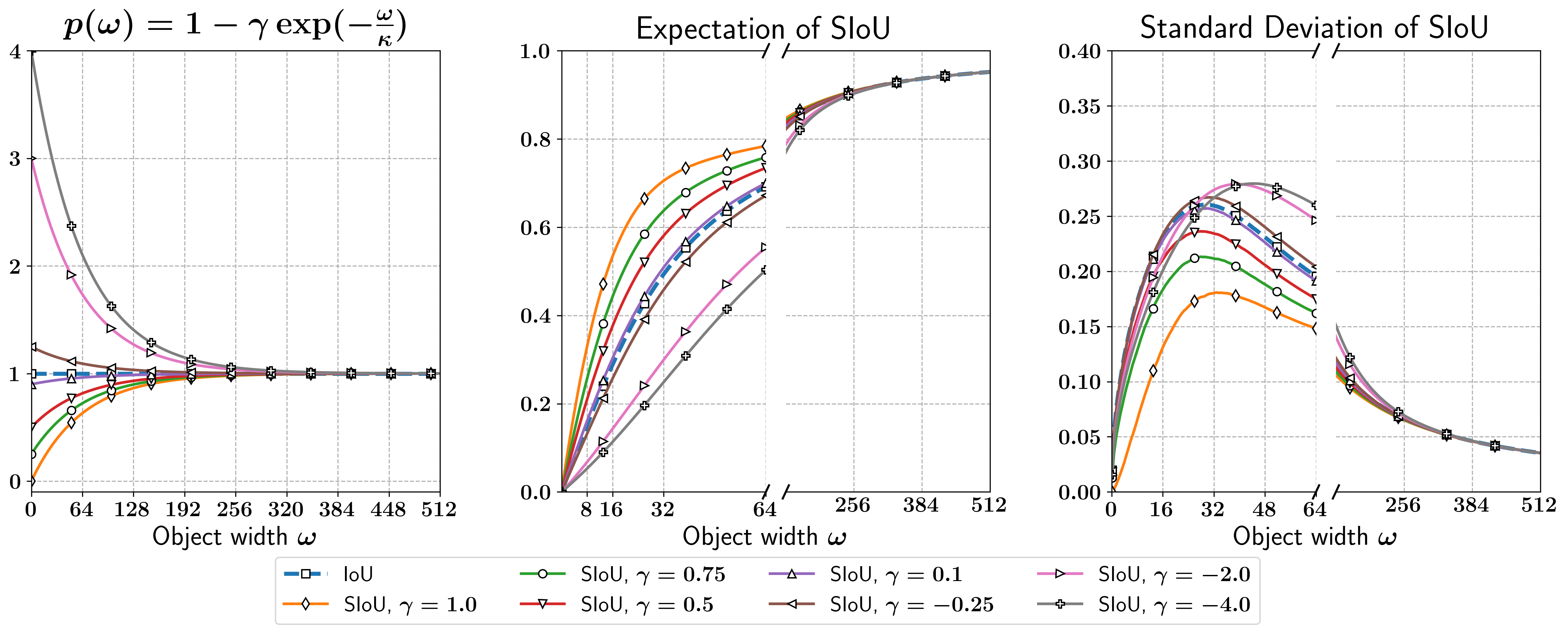}
       \caption{$\gamma$'s influence, with $\kappa=64$}
       \label{fig:gamma_influence}
   \end{subfigure}%
   
   \begin{subfigure}[b]{\textwidth}
       \centering
       \includegraphics[width=0.93\textwidth]{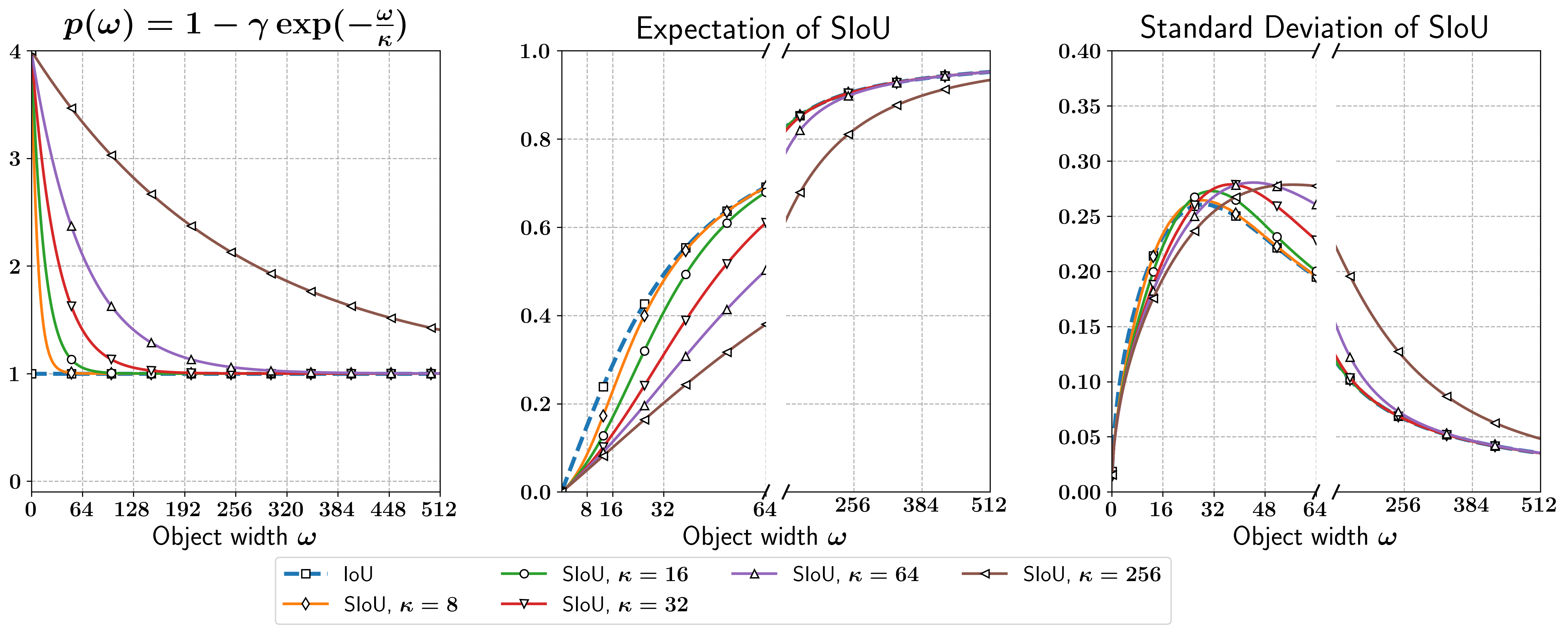}
       \caption{$\kappa$'s influence, with $\gamma=-3$}
       \label{fig:kappa_influence}
   \end{subfigure}
   \caption{Influence of $\gamma$ and $\kappa$ on the expected value and standard deviation of SIoU.}
   \label{fig:influence_gamma_kappa}
\end{figure*}

\noindent
\textbf{$\gamma$ and $\kappa$ influence on SIoU}\\
First, following the analysis from \cref{sec:empirical_analysis},
\cref{fig:influence_gamma_kappa} investigates the influence of $\gamma$ and
$\kappa$ on SIoU behavior. \cref{fig:gamma_influence} shows the value of $p$,
the expectation and variance of SIoU against object size for $\gamma \in \{1.0,
0.75, 0.5, 0.1, 0.0, -0.25, -2, -4\}$. $p$ is a function of the average area of
the boxes ($\tau = \frac{w_1h_1 + w_2h_2}{2})$, for simplicity we suppose here
that the boxes are squares of the same width $\omega$, hence $\sqrt{\tau} =
\omega$. Then $p$ can be viewed as a function of $\omega$: $p(\omega) = 1 -
\gamma\exp(-\omega / \kappa)$. For negative values of $\gamma$, $p$
decreases from $p(0) = 1 - \gamma$ to 1, small objects get higher exponents in
comparison with larger objects. On the contrary, when $\gamma > 0$, $p$
increases from $p(0) = 1 - \gamma$ to 1. Changing $\gamma$ also influences the
distribution of SIoU. As $\gamma$ increases, the expected value for small
objects increases as well, while the variance decreases. 

\cref{fig:kappa_influence} shows the same curves for $\kappa \in
\{8,16,32,64,256\}$. $\kappa$ controls how fast $p$ approaches 1 and
therefore, changing $\kappa$ simply shifts the curves of expectation and
variance accordingly. As $\kappa$ increases, IoU's behavior is retrieved for
larger objects reducing the expected value of SIoU. The variance is not changed
much by $\kappa$, it slightly shifts the maximum of the curve, \ie the object
size for which SIoU's variance is maximum.

\begin{figure*}
   \centering
   \includegraphics[width=\textwidth]{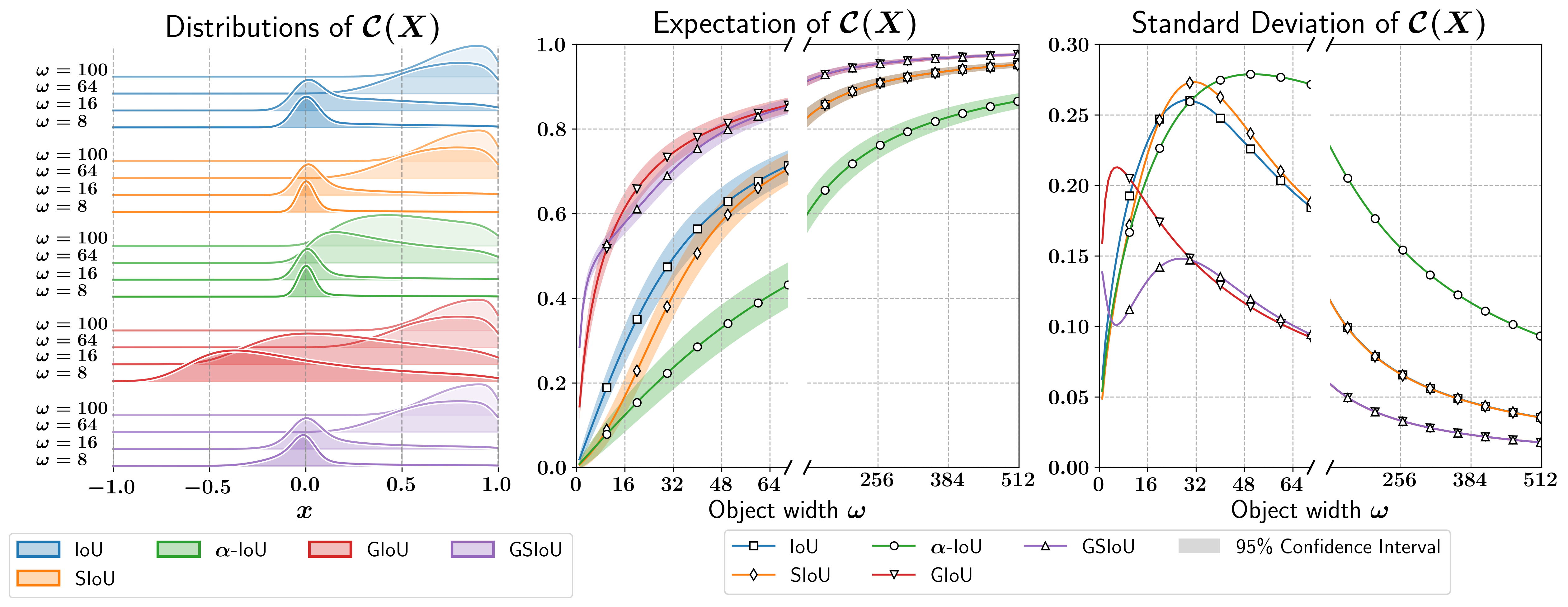}
   \vspace{0.5mm}
   \caption{Analysis of the distribution of IoU, SIoU, GIoU, GSIoU and
           $\alpha$-IoU when computed on inaccurately positioned boxes. This is
           done by observing the probability distribution functions (pdfs) for various
           $\omega$ values \textbf{(left)}, the expectation \textbf{(middle)}
           and standard deviation \textbf{(right)} for all criteria. For SIoU
           and GSIoU, we fixed $\gamma=-4$ and $\kappa=16$, for $\alpha$-IoU,
           $\alpha=3$ (as recommended in the original paper
           \cite{he2021alpha}). The inaccuracy of the detector is set to
           $\sigma=16$. Note that the empirical pdfs were smoothed using a
           Kernel Density Estimator method. This affects particularly IoU, SIoU
           and $\alpha$-IoU for the actual pdf is defined only on $[0,1]$. For
           the sake of visualization, GIoU and GSIoU were rescaled between 0
           and 1 for the expectation and standard deviation plots. }
   \label{fig:criteria_analysis_4}
\end{figure*}

\cref{fig:criteria_analysis_4} also provides pdfs plots for various object
sizes for SIoU and GSIoU, in addition to expectation and variance comparison
between existing criteria. For this figure, $\gamma=-3$ and $\kappa=16$. This
figure echoes \cref{fig:criteria_analysis} which plots the same curves but with
$\gamma=0.5$ and $\kappa=64$. 

\noindent
\textbf{$\gamma$ and $\kappa$ influence on FSOD Performance} \\
The setting of $\gamma$ and $\kappa$ are crucial for training. Therefore, we
conducted various experiments on dataset DOTA to find the best parameters
for GSIoU loss. The results can be found in
\cref{tab:gamma_perf,tab:kappa_perf}. This shows that the optimal values for
DOTA dataset are $\gamma=-3$ and $\kappa=8$. However, $\kappa=16$ is also a good
choice and seems more consistent across datasets. Thus, we choose to keep
$\gamma=-3$ and $\kappa=16$ for other experiments. An exhaustive grid search should
be done to find even better settings. Our search was sparse and a better
combination of $\gamma$ and $\kappa$ probably exists. 

\begin{table}[h]
    \centering
    \resizebox{\columnwidth}{!}{%
    \begin{tabular}{@{}ccccccccc@{}}
        \toprule[1.5pt]
        \multicolumn{1}{l}{}      & \multicolumn{4}{c}{\textbf{Base classes}}                         & \multicolumn{4}{c}{\textbf{Novel Classes}}                                                                 \\ \midrule
        $\bm{\gamma}$ & \textbf{All}   & \textbf{S}     & \textbf{M}     & \textbf{L}     & \textbf{All}   & \textbf{S}                  & \textbf{M}                   & \textbf{L}                   \\
        \textbf{0.5}                       & 47.09          & 21.29          & 54.67          & 65.48          & 30.50          & 8.83                        &44.97                         & 65.89 \\
        \textbf{0.25}                      & 45.94          & 21.60          & 54.39          & 63.40          & 30.96          & 12.53                       & 42.37                        & 64.14                        \\
        \textbf{0}                         & 52.41          & 26.94          & 61.17          & 63.00          & 41.03          & 24.01                       & 52.13                        & 69.78                        \\
        \textbf{-0.5}                      & 52.80          & \textbf{27.16} & 61.19          & 64.61          & 41.06          & 25.20                       & 50.18                        & \textbf{72.04}               \\
        \textbf{-1}                        & 53.03          & 23.20          & 61.53          & 66.68          & 42.77          & 27.55                       & \textbf{52.01}               & 70.76                        \\
        \textbf{-2}                        & \textbf{54.06} & 23.68          & \textbf{62.69} & 66.62          & 43.67          & 30.04                       & 51.69                        & 69.66                        \\
        \textbf{-3}                        & 52.91          & 22.14          & 61.19          & 66.02          & \textbf{45.88} & \textbf{34.83}              & 51.26                        & 70.78                        \\
        \textbf{-4}                        & 53.59          & 22.50          & 62.48          & 66.18          & 42.43          & 27.56                       & 51.79                        & 68.70                        \\
        \textbf{-9}                        & 53.11          & 20.98          & 62.13          & \textbf{67.00} & 42.63          & 30.53                       & 48.89                        & 68.62                        \\ \bottomrule[1.5pt]
        \end{tabular}%
    }
    \caption{Evolution of the few-shot performance (XQSA with GSIoU loss) on DOTA for various values of $\gamma$ ($\kappa=16$ is fixed).}
    \label{tab:gamma_perf}
    \end{table}

 \begin{table}[h]
    \centering
    \resizebox{\columnwidth}{!}{%
    \begin{tabular}{@{}ccccccccc@{}}
    \toprule[1.5pt]
    \multicolumn{1}{l}{\textbf{}} & \multicolumn{4}{c}{\textbf{Base classes}}           & \multicolumn{4}{c}{\textbf{Novel Classes}}          \\ \midrule
    \textbf{$\boldsymbol \kappa$}                & \textbf{All} & \textbf{S} & \textbf{M} & \textbf{L} & \textbf{All} & \textbf{S} & \textbf{M} & \textbf{L} \\
    \textbf{4}                    & 51.65        & 21.50      & 59.76      & 65.85      & 42.98        & 30.33      & 48.57      & \textbf{73.41}      \\
    \textbf{8}                    & 52.70        & 21.96      & 61.49      & 66.43      & \textbf{44.16}        & \textbf{31.35}      & 50.70      & 71.99      \\
    \textbf{16}                   & \textbf{54.06}        & \textbf{23.68}      & 62.69      & 66.62      & 43.67        & 30.04      & 51.69      & 69.66      \\
    \textbf{32}                   & 53.88        & 22.33      & \textbf{63.00}      & \textbf{67.35}      & 37.36        & 23.65      & 44.60      & 66.29      \\
    \textbf{64}                   & 52.82        & 21.79      & 61.46      & 66.77      & 43.68        & 29.43      & \textbf{52.47}      & 69.46      \\
    \textbf{128}                  & 53.42        & 21.73      & 62.90      & 66.75      & 41.32        & 26.85      & 49.40      & 70.38      \\ \bottomrule[1.5pt]
    \end{tabular}%
    }
    \caption{Evolution of the few-shot performance (XQSA with GSIoU loss) for various values of $\kappa$ ($\gamma= -2$ is fixed).}
    \label{tab:kappa_perf}
    \end{table}

\noindent
\textbf{Influence of $\boldsymbol \gamma$ and $\boldsymbol \kappa$ on FSOD performance on Pascal VOC dataset} \\
The search conducted above was carried out on DOTA dataset. It transposes nicely
on DIOR dataset as well. Yet, these two datasets are similar. They
both contain aerial images with some classes in common, but most importantly,
the size of their objects are highly similar. When applied to different
datasets, these results may not hold. For instance, Pascal VOC requires other
combinations of $\gamma$ and $\kappa$ to outperform the training with GIoU. This is
shown in \cref{tab:pascal_search}.

\begin{table}[h]
    \centering
    \resizebox{\columnwidth}{!}{%
    \begin{tabular}{@{}ccccccccc@{}}
    \toprule[1.5pt]
                                    & \multicolumn{4}{c}{\textbf{Base Classes}} & \multicolumn{4}{c}{\textbf{Novel Classes}} \\ \midrule
    \textbf{Loss}                   & \textbf{All}    & \textbf{S}      & \textbf{M}      & \textbf{L}     & \textbf{All}    & \textbf{S}      & \textbf{M}      & \textbf{L}      \\
    \textbf{GIoU}                           & 51.09  & \textbf{13.93}  & 40.26  & 62.01 & 48.42  & 18.44  & 36.06  & 59.99  \\
    \textbf{GSIoU} $\boldsymbol \gamma=-3$, $\boldsymbol \kappa=16$  & 45.22  & 10.06  & 34.85  & 57.10 & 43.16  & 14.89  & 33.92  & 54.16  \\
    \textbf{GSIoU} $\boldsymbol \gamma=-1$, $\boldsymbol \kappa=64$  & 54.47  & 13.88  & 40.13  & 66.82 & 55.16  & \textbf{22.94}  & 36.24  & 67.40  \\
    \textbf{GSIoU} $\boldsymbol \gamma=0.5$, $\boldsymbol \kappa=64$ & \textbf{56.97}  & 13.88  & \textbf{40.75}  & \textbf{70.31} & \textbf{55.36}  & 20.25  & \textbf{36.85}  & \textbf{68.05}  \\ \bottomrule[1.5pt]
    \end{tabular}
    }
    \caption{Few-shot performance on Pascal VOC dataset with different values of $\gamma$ and $\kappa$.}
    \label{tab:pascal_search}
    \end{table}

\section{Proofs of Properties}
\label{app:properties}

In this appendix, we provide the proofs for
\cref{property:relaxation,property:l_g_reweight}, and discuss the
\textit{order-preservigness} of SIoU.

\begin{appproperty}[SIoU Relaxation]
    \label{app_property:app_relaxation}
    Let $b_1$ and $b_2$ be two bounding boxes and introduce $\tau = \frac{w_1h_1 +
    w_2h_2}{2}$ their average area. SIoU preserves the behavior of IoU in
    certain cases such as:
   \begin{itemize}
    \setlength\itemsep{0em}
        \item[-]$\textup{IoU}(b_1, b_2) = 0 \Rightarrow \textup{SIoU}(b_1, b_2) =\textup{IoU}(b_1, b_2) = 0$
        \item[-]$\textup{IoU}(b_1, b_2) = 1 \Rightarrow\textup{SIoU}(b_1, b_2) =\textup{IoU}(b_1, b_2) = 1$
        \item[-]$\lim\limits_{\tau \to +\infty} \textup{SIoU}(b_1, b_2) =\textup{IoU}(b_1,b_2)$
        \item[-]$\lim\limits_{\kappa\to 0} \textup{SIoU}(b_1, b_2) =\textup{IoU}(b_1,b_2)$
   \end{itemize}
\end{appproperty}

\begin{proof}
    First let recall the expression of SIoU, $\text{SIoU}(b_1, b_2) =
    \text{IoU}(b_1, b_2) ^ p$ with $p = 1 - \gamma
    \exp\left({-\frac{\sqrt{\tau}}{\kappa}}\right)$. $\tau >0$ because boxes
    cannot be empty and as $\gamma \in ]-\infty, 1]$ and $\kappa \in
    \mathbb{R}_+^*$, we have $p>0$.\\

    \vspace{5mm}
    \noindent
    From this, the first two items of \cref*{property:relaxation} follow clearly.\\ 

    \noindent
    The two other points hold because the function $f \colon x \mapsto
    \text{IoU}(b_1, b_2)^x $ is continuous on $\mathbb{R}$ for any couple of
    boxes $b_1$ and $b_2$ ($\text{IoU}(b_1, b_2) \in [0,1]$) and  
    $\lim\limits_{\tau \to \infty} p = \lim\limits_{\kappa \to 0} p = 1$.

\end{proof}

\setcounter{equation}{3}
\begin{appproperty}[Loss and gradients reweighting]
    \label{property:app_l_g_reweight}
    Let $\mathcal{L}_{\textup{IoU}}(b_1, b_2) = 1 - \textup{IoU}(b_1, b_2)$ and
    $\mathcal{L}_{\textup{SIoU}}(b_1, b_2) = 1 - \textup{SIoU}(b_1, b_2)$ be the
    loss functions associated respectively with IoU and SIoU. Let denote the
    ratio between SIoU and IoU losses by 
    $\mathcal{W}_{\mathcal{L}}(b_1, b_2) =
    \frac{\mathcal{L}_{\textup{SIoU}}(b_1, b_2)}{\mathcal{L}_{\textup{IoU}}(b_1,
    b_2)}$. 
    Similarly, 
    $\mathcal{W}_{\mathcal{\nabla}}(b_1, b_2) =
    \frac{|\nabla\mathcal{L}_{\textup{SIoU}}(b_1, b_2)|}{|\nabla\mathcal{L}_{\textup{IoU}}(b_1,
    b_2)|}$ 
    denotes the ratio of gradients generated from SIoU and IoU losses: 
    \begin{align}
        \mathcal{W}_{\mathcal{L}}(b_1, b_2) &= \frac{1- \textup{IoU}(b_1, b_2)^p}{1-\textup{IoU}(b_1, b_2)}, \\
        \mathcal{W}_{\mathcal{\nabla}}(b_1, b_2) &= p\textup{IoU}(b_1, b_2)^{p-1},
    \end{align}

    \noindent
    $\mathcal{W}_{\mathcal{L}}$ and $\mathcal{W}_{\mathcal{\nabla}}$ are
    increasing (resp. decreasing) functions of IoU when $p\geq 1$ (resp. $p <
    1$) which is satisfied when $\gamma \leq 0$ (resp. $\gamma > 0$). As the IoU
    goes to 1, $\mathcal{W}_{\mathcal{L}}$ and $\mathcal{W}_{\mathcal{\nabla}}$
    approaches $p$: 
    \begin{align}
        \lim\limits_{\textup{IoU}(b_1, b_2) \to 1}\mathcal{W}_{\mathcal{L}}(b_1, b_2) &=  p, \\
        \lim\limits_{\textup{IoU}(b_1, b_2) \to 1}\mathcal{W}_{\mathcal{\nabla}}(b_1, b_2) &=  p.
    \end{align}

\end{appproperty}

\begin{proof}
    Let denote the $\text{IoU}(b_1, b_2)$ by $\mu$, and define two functions $ f
    \colon \mu \mapsto 1- \mu = \mathcal{L}_{\text{IoU}}(b_1, b_2)$ and $g
    \colon \mu \mapsto 1- \mu^p = \mathcal{L}_{\text{SIoU}}(b_1, b_2)$. \\
    
    \noindent
    $f$ and $g$ are differentiable on $[0,1]$ and $\lim\limits_{\mu \to 1}
    f(\mu) = \lim\limits_{\mu \to 1} g(\mu) = 0$. This holds because $p$ is
    independent of the IoU (\ie $\mu$).
    
    \noindent
    Therefore L'Hôpital's rule can be applied:\\
    $\lim\limits_{\mu \to 1}\mathcal{W}_{\mathcal{L}} = \lim\limits_{\mu \to
    1}\frac{g(\mu)}{f(\mu)} = \lim\limits_{\mu \to 1}\frac{g'(\mu)}{f'(\mu)} =
    \lim\limits_{\mu \to 1} p\mu^{p-1} = p$.
    
    \noindent
    The expression of the second ratio $\mathcal{W}_{\mathcal{\nabla}}(b_1,
    b_2)$ follows directly as $|\nabla\mathcal{L}_{\textup{SIoU}}(b_1, b_2)| =
    g'(\mu)$ and $|\nabla\mathcal{L}_{\textup{IoU}}(b_1, b_2)| = f'(\mu)$, hence
    $\lim\limits_{\mu \to 1}\mathcal{W}_{\nabla} = \lim\limits_{\mu \to 1}
    p\mu^{p-1} = p$.
\end{proof}

\noindent
\textbf{Order-preservigness} \\
Let us take three boxes $b_1$, $b_2$, and $b_3$. Order-preservigness does not
hold for SIoU. Therefore $\text{IoU}(b_1, b_2) \leq  \text{IoU}(b_1, b_3)$ does
not imply $\text{SIoU}(b_1, b_2) \leq  \text{SIoU}(b_1, b_3)$. However, this
property is often true in practice. Denoting by $\tau_{i,j}$ the average area
between boxes $i$ and $j$, we can study the conditions for the order to hold. We
will also note $p_{i,j} = 1- \gamma
\exp\left({-\frac{\sqrt{\tau_{i,j}}}{\kappa}}\right)$\\

\noindent
Let's suppose, without loss of generality, that $\tau_{1,2} \leq \tau_{1,3}$.
Otherwise, cases 1 and 2 would be swapped. \\

\noindent
\textbf{Case 1: } $\gamma \leq 0$\\
We have $p_{1,2} > p_{1,3}$ as $\tau_{1,2} \leq \tau_{1,3}$ and $\gamma \leq 0$.
Therefore, $p_{1,2} = p_{1,3} + \varepsilon$, with $\varepsilon > 0$. \\

\noindent
Then,
\begin{equation*}
    \begin{aligned}
        \text{IoU}(b_1,b_2)^{p_{1,2}} &= \text{IoU}(b_1,b_2)^{p_{1,3} +\varepsilon}\\
                                        & = \text{IoU}(b_1,b_2)^{p_{1,3}}\text{IoU}(b_1,b_2)^{\varepsilon}\\
                                        &\leq \text{IoU}(b_1,b_2)^{p_{1,3}} \\
                                        &\leq \text{IoU}(b_1,b_3)^{p_{1,3}}
    \end{aligned}
\end{equation*}

\noindent
Line 3 holds because $0 < \text{IoU}(b_1,b_2)^{\varepsilon} \leq 1$. Line 4 is
true because $\text{IoU}(b_1, b_2) \leq  \text{IoU}(b_1, b_3)$ and the function
$h\colon x \mapsto x^{p_{1,3}}$ is monotonically increasing. Hence, when $\gamma
\leq 0$ the order is preserved.\\

\noindent
\textbf{Case 2: } $\gamma > 0$\\
We have $p_{1,3} > p_{1,2}$ as $\tau_{1,3} \leq \tau_{1,2}$ and $\gamma > 0$.
Therefore, $p_{1,3} = p_{1,2} + \varepsilon$, with $\varepsilon > 0$. \\

\noindent
In this case, the order does not always hold, counter-examples can be found.
However, it is useful to study the conditions for it to hold:
$$\text{IoU}(b_1, b_2)^{p_{1,2}} \leq  \text{IoU}(b_1, b_3)^{p_{1,3}} \Leftrightarrow \frac{\ln
\big(\text{IoU}(b_1, b_2)\big)}{\ln\big(\text{IoU}(b_1, b_3)\big)} \geq \frac{p_{1,3}}{p_{1,2}}$$

\noindent
In practice, the right condition is often true as $p_{1,2}$ and $p_{1,3}$ are
close due to scale matching, a trick present in most detection frameworks to
prevent comparison of proposals and ground truth of very different sizes. In
addition, the ratio of log values gets large rapidly, even if the gap between
$\text{IoU}(b_1, b_2)$ and $\text{IoU}(b_1, b_3)$ is small, the ratio $\frac{\ln
\big(\text{IoU}(b_1, b_2)\big)}{\ln\big(\text{IoU}(b_1, b_3)\big)}$ can be
large.\\

\noindent
The order-preservigness property holds in many cases. This is sensible as
IoU is still a reliable metric that has been used extensively for the training
and evaluation of detection models. However, in the rare cases where this order
is broken, the $\text{IoU}(b_1, b_2)$ and $\text{IoU}(b_1, b_3)$ are close, so
IoU does not discriminate much between the boxes. On the contrary, SIoU prefers
the smallest one (or largest one, according to the choice of $\gamma$). This
stronger discrimination is probably beneficial for training and evaluation.

\section{Inaccuracy Analysis}
\label{app:inaccuracy}

\subsection{Shift plots for GIoU and GSIoU}
\begin{figure}[h]
    \centering
    \includegraphics[width=\columnwidth]{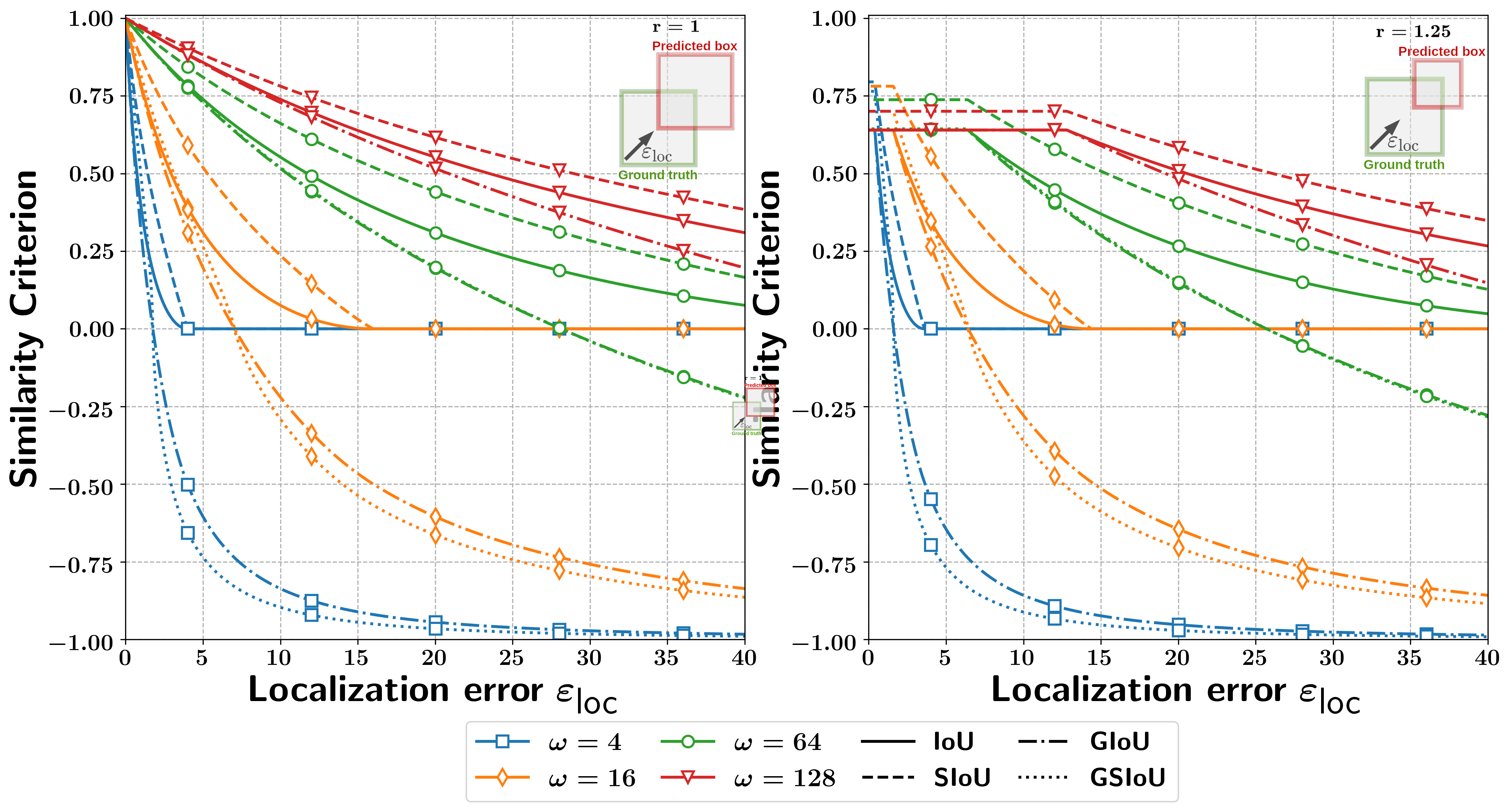}
    \caption{Evolution of various criteria (IoU, GIoU, and GSIoU)
    when a box is shifted from the ground truth box by $\rho$ pixels for various
    box sizes $\omega \in \{4, 16, 64, 128\}$. With boxes of the same size
    \textbf{(left)} and different sizes \textbf{(right)}.}
    \label{fig:shift_gsiou}
\end{figure}

\subsection{Inaccuracy Tolerance Assumptions}
Several assumptions were made in \cref{sec:inaccuracy} to analyze the
criteria for box similarity:
\begin{enumerate}[noitemsep]
    \item Boxes are shifted only horizontally. 
    \item Boxes have the same size.
    \item The detector's inaccuracy is fixed and does not depend on the object size.
\end{enumerate}

\begin{figure}
    \centering
    \begin{subfigure}[t]{\columnwidth}
        \centering
        \includegraphics[trim=490 0 0 0, clip,width=\columnwidth]{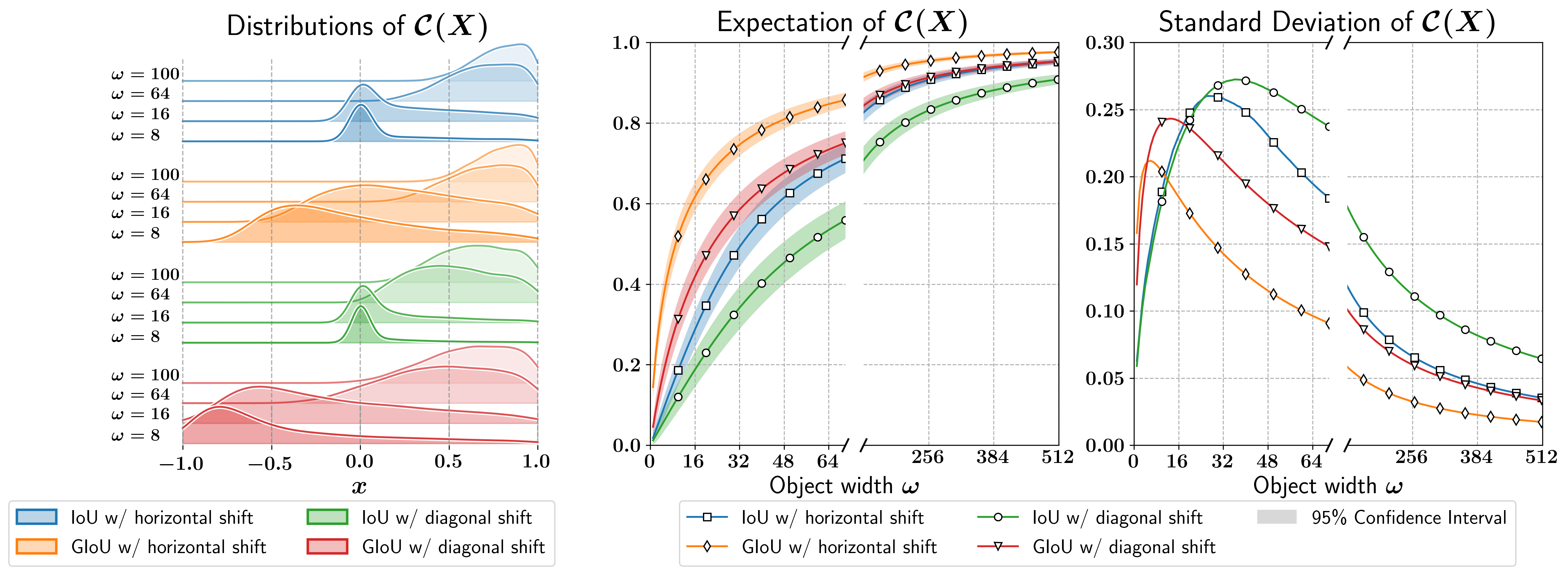}
        \caption{IoU and GIoU expected values and standard deviation with horizontally and diagonally shifted boxes.}
        \label{fig:diagonal_shift_distri}
    \end{subfigure}%
    
    \begin{subfigure}[b]{\columnwidth}
        \centering
        \includegraphics[trim=430 0 0 0, clip,width=\columnwidth]{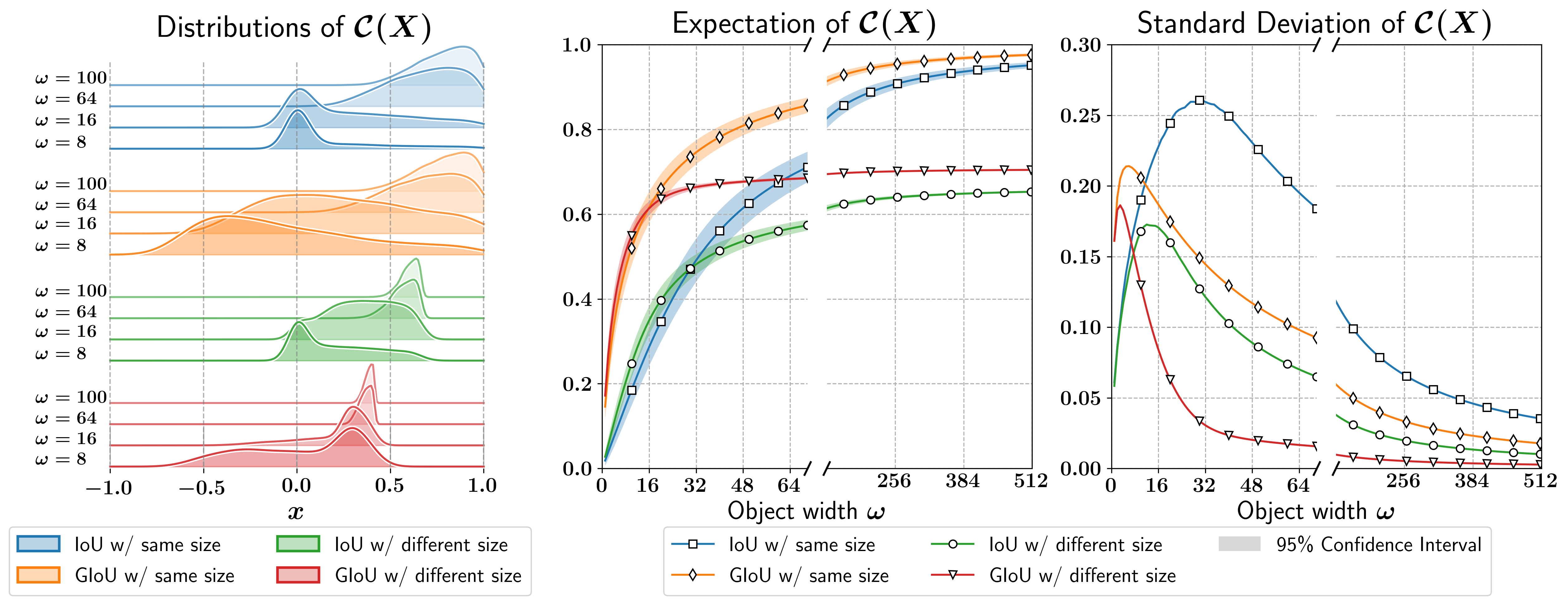}
        \caption{IoU and GIoU expected values and standard deviation with and without boxes of the same size.}
        \label{fig:different_size_boxes}
    \end{subfigure}
    \begin{subfigure}[b]{\columnwidth}
        \centering
        \includegraphics[trim=490 0 0 0, clip,width=\columnwidth]{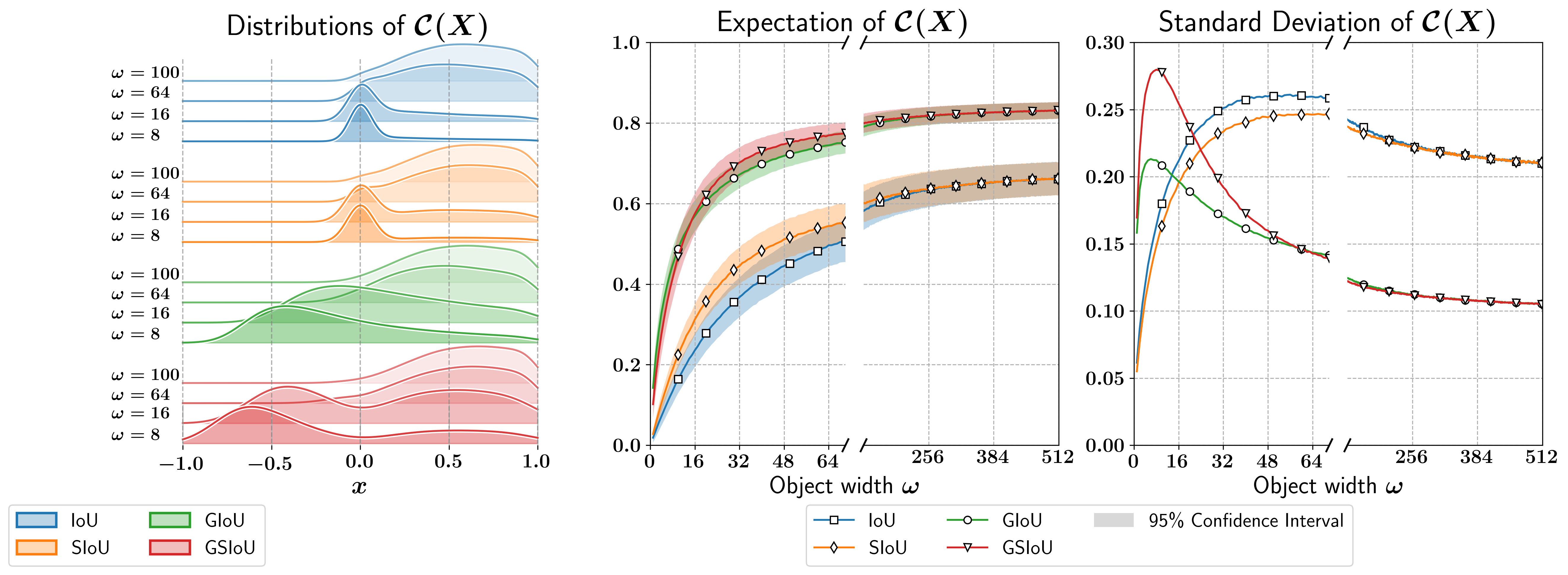}
        \caption{IoU, SIoU, GIoU, and GSIoU expected values and standard
        deviation with the detector's inaccuracy modeled as an affine function,
        $\sigma(\omega) = \sigma_0 + \lambda \omega$ ($ \sigma_0 = 16$,
        $\lambda=\frac{1}{4}$).}
        \label{fig:affine_inaccuracy}
    \end{subfigure}
    \caption{Relaxing the constraints for criteria' distribution analysis.}
    \label{fig:relax_assumptions}
\end{figure}

The first two assumptions are relatively harmless. Allowing diagonal shifts
simply accelerates the IoU decreasing rate. A 1-pixel diagonal shift is
equivalent to a vertical and a horizontal shift. Intuitively, this is similar to
a 2-pixels horizontal shift. This is not true because, with a 1-pixel diagonal
shift, the area of intersection decreases slower than with a 2-pixels horizontal
shift. Following the notations from \cref{sec:inaccuracy}, the intersection
between two boxes of width $\omega$ diagonally shifted by $\rho$ pixels is
$(\omega - \rho)^2 = \omega^2 - 2\omega\rho + \rho^2$ while the intersection
between same boxes horizontally shifted by $2\rho$ pixels is $\omega(\omega -
2\rho) = \omega^2 - 2\omega\rho$. To ensure that this does not question the
conclusions of \cref{sec:empirical_analysis}, \cref{fig:diagonal_shift_distri}
compares the expected values and variances of IoU and GIoU with horizontal
and diagonal shifts. Similar behaviors are observed with or without diagonal
shifting. The only difference is that the expected values of the criteria for
diagonally shifted boxes are lower as the shifts get larger. It also increases
the variances as the distributions are more spread. Relaxing the second
constraint also results in slightly different distributions, but with similar
behavior. Having different size boxes only changes the maximum value of the
criteria. If boxes have different sizes, the maximum value must be smaller than
1. Therefore, the expected values approach smaller values than 1 as objects get
larger. And the variance is reduced as the range of criteria values is smaller.
This is visible in \cref{fig:different_size_boxes}. 

It is less straightforward that the conclusions hold without the last
assumption. In the analysis from \cref{sec:empirical_analysis}, we assume that
the inaccuracy of the detector is fixed. This means that the detector generates
randomly shifted boxes by the same number of pixels on average no matter the
size of the object. This is certainly false, in practice, in terms of absolute
distance, detectors are better with smaller objects. However, the inaccuracy
cannot simply be proportional to object sizes because small objects would then
be perfectly detected. Thus, we tried to change the inaccuracy of the detector
as an affine function of the box width: $\sigma(\omega) = \sigma_0 + \lambda
\omega$. We choose to set $\sigma_0$ to the fixed value of $\sigma$ used in
\cref{sec:empirical_analysis} and $\lambda=1/4$. This setting better reflects
the inaccuracy of a true detector. The expected values and standard
deviations of IoU, SIoU, GIoU, and GSIoU with this inaccuracy setting are plotted
in \cref{fig:affine_inaccuracy}. The main difference with fixed inaccuracy is
that expected values do not approach 1 as object size gets larger, instead they
tend towards lower values. It also leads to non-zero variance for
large objects. For small objects, however, the curves of the different criteria
are mostly unchanged, and the conclusions formulated in
\cref{sec:empirical_analysis} are still valid.\\

\subsection{Theoretical GIoU's distribution}
\noindent
In this appendix, we give a proof of the formulas of the probability density
function, expected value, and variance of GIoU. We also provide some
non-closed-form expressions for other criteria.
\begin{appproposition}[GIoU's distribution]
    \label{prop:appdistribution_giou}
    Let $b_1 = (0,y_1,w_1,h_1)$ be a bounding box horizontally centered and $b_2
    = (X,y_2,w_2,h_2)$ another bounding box randomly positioned, with $X\sim
    \mathcal{N}(0, \sigma^2)$ and $\sigma \in \mathbb{R}^*_+$. Let's suppose that
    the boxes are identical squares, shifted only horizontally (\ie $w_1 = w_2 = h_1 =
    h_2$ and $y_1 = y_2$). \\
    \noindent
    Let $Z = \mathcal{C}(X)$, where $\mathcal{C}$ is the generalized intersection over union. The
    probability density function of $Z$ is given by:
    \begin{align}
        d_Z(z) &= \frac{4\omega}{(1+z)^2\sqrt{2\pi} \sigma} \exp\left(-\frac{1}{2}\left[\frac{\omega(1-z)}{\sigma(1+z)}\right]^2\right).
    \end{align}

    \noindent
    The two first moments of $Z$ exist and are given by:
    \begin{align}
        \mathbb{E}[Z] &= \frac{2}{\pi^{3/2}} G^{2,3}_{3,2}\left(2a^2 \bigg\rvert \begin{matrix}0 & \frac{1}{2} & \frac{1}{2} \\ \frac{1}{2}& 0 \end{matrix}\right), \\
        \mathbb{E}[Z^2] &= 1- \frac{8a}{\sqrt{2\pi}} + \frac{16a^2}{\pi^{3/2}} G^{2,3}_{3,2}\left(2a^2 \bigg\rvert \begin{matrix}-1 & \frac{1}{2} & -\frac{1}{2} \\ \frac{1}{2}  & 0 \end{matrix}\right),
    \end{align}

    \noindent
    where $G$ is the Meijer G-function \cite{meijer2013} and $a=\frac{\sigma}{\omega}$. 
\end{appproposition}

\begin{table*}[]
    \centering
    \resizebox{\textwidth}{!}{%
    \begin{tabular}{@{}ccccccccccccccccc@{}}
    \toprule[1.5pt]
                    & \textbf{Criterion Expression} & \textbf{Probability Density Function} & $\boldsymbol{\mathbb{E}[Z]}$ & $\boldsymbol{\mathbb{E}[Z^2]}$ \\ \midrule[1.5pt]
    \textbf{IoU}             & \parbox{0.25\textwidth}{\begin{align*}\mathcal{C} \colon \mathbb{R} &\to [0,1] \\x &\mapsto \max\left(0, \frac{\omega- |x|}{\omega + |x|}\right)\end{align*}} &\parbox{0.25\textwidth}{\begin{align*}d_Z(z) = 2\bigg[(&1 - F_X(\omega\frac{1-z}{1+z}) \delta_0(z) \\&+ \mathds{1}_{\mathbb{R}_+}(z) \frac{4\omega}{(1+z)^2} d_X(\omega\frac{1-z}{1+z}))\bigg]\end{align*}  } &     \parbox{0.25\textwidth}{\begin{align*}\mathbb{E}[Z] = \frac{4}{\sqrt{2\pi}a}\int_{0}^{1} \frac{1}{1+u}e^{-\frac{u^2}{2a^2}}\,du  - \text{erf}(\frac{1}{\sqrt{2}a})\\\end{align*}} &     \parbox{0.25\textwidth}{\begin{align*}\mathbb{E}[Z^2]= \text{erf}(\frac{1}{\sqrt{2}a}) - \frac{8a}{\sqrt{2\pi}a}\bigg[&1- \frac{1}{2}e^{-\frac{1}{2a^2}}\\&-2\int_{0}^{1} \frac{1}{(1+u)^3}e^{-\frac{u^2}{2a^2}}\,du \bigg]\end{align*}} &\\\bottomrule[1.5pt]
    \textbf{GIoU}           & \parbox{0.25\textwidth}{\begin{align*}\mathcal{C} \colon \mathbb{R} &\to [-1,1] \\x &\mapsto \frac{\omega- |x|}{\omega + |x|}\end{align*}} &\parbox{0.25\textwidth}{\begin{align*}d_Z(z) &= \frac{4\omega}{(1+z)^2\sqrt{2\pi} \sigma} \exp\left(-\frac{1}{2}\left[\frac{\omega(1-z)}{\sigma(1+z)}\right]^2\right)\end{align*}} &     \parbox{0.25\textwidth}{\begin{align*}\mathbb{E}[Z] &= \frac{2}{\pi^{3/2}} G^{2,3}_{3,2}\left(2a^2 \bigg\rvert \begin{matrix}0 & \frac{1}{2} & \frac{1}{2} \\ \frac{1}{2}& 0 \end{matrix}\right) \\\end{align*}} &     \parbox{0.25\textwidth}{\begin{align*}\mathbb{E}[Z^2] &= 1- \frac{8a}{\sqrt{2\pi}} + \frac{16a^2}{\pi^{3/2}} G^{2,3}_{3,2}\left(2a^2 \bigg\rvert \begin{matrix}-1 & \frac{1}{2} & -\frac{1}{2} \\ \frac{1}{2}  & 0 \end{matrix}\right)\end{align*}} &\\\bottomrule[1.5pt]
    \textbf{SIoU}            & \parbox{0.25\textwidth}{\begin{align*}\mathcal{C} \colon \mathbb{R} &\to [0,1] \\x &\mapsto \begin{cases} \left(\frac{\omega- |x|}{\omega + |x|}\right)^p \,&\text{if  } \omega- |x| \geq 0\\ 0\,&\text{otherwise}\end{cases}\end{align*}} &\parbox{0.25\textwidth}{\begin{align*}d_Z(z) = 2\bigg[(&1 - F_X(\omega\frac{1-z^{1/p}}{1+z^{1/p}}) \delta_0(z) \\&+ \mathds{1}_{\mathbb{R}_+}(z) \frac{4\omega z^{1/p -1 }}{(1+z^{1/p})^2} d_X(\omega\frac{1-z^{1/p}}{1+z^{1/p}}))\bigg]\end{align*}} &     \parbox{0.25\textwidth}{\begin{align*}\mathbb{E}[Z] &= 2\omega\int_{0}^{1} \big(\frac{1-u}{1+u}\big)^p e^{-\frac{u^2}{2a^2}}\,du\end{align*}} &     \parbox{0.25\textwidth}{\begin{align*}\mathbb{E}[Z^2] &= 2\omega\int_{0}^{1} \big(\frac{1-u}{1+u}\big)^{2p} e^{-\frac{u^2}{2a^2}}\,du\end{align*}} &\\\bottomrule[1.5pt]
    \textbf{GSIoU}            & \parbox{0.25\textwidth}{\begin{align*}\mathcal{C} \colon \mathbb{R} &\to [-1,1] \\x &\mapsto \begin{cases} \left(\frac{\omega- |x|}{\omega + |x|}\right)^p \,&\text{if  } \omega- |x| \geq 0\\ -\left(\frac{|x| - \omega}{\omega + |x|}\right)^p \,&\text{otherwise}\end{cases}\end{align*}} &\parbox{0.25\textwidth}{\begin{align*}d_Z(z) = 2\bigg[(&1 - F_X(\omega\frac{1-z^{1/p}}{1+z^{1/p}}) \delta_0(z) \\&+ \mathds{1}_{\mathbb{R}_+}(z) \frac{4\omega z^{1/p -1 }}{(1+z^{1/p})^2} d_X(\omega\frac{1-z^{1/p}}{1+z^{1/p}}))\\&-( 1 - F_X(\omega\frac{1+|z|^{1/p}}{1-|z|^{1/p}}) \delta_0(z)) \\&+ \mathds{1}_{\mathbb{R}_-}(z) \frac{4\omega |z|^{1/p -1 }}{(1-|z|^{1/p})^2} d_X(\omega\frac{1+|z|^{1/p}}{1-|z|^{1/p}}))\bigg]\end{align*}} &     \parbox{0.25\textwidth}{\begin{align*}\mathbb{E}[Z] = 2\omega\bigg[&\int_{0}^{1} \big(\frac{1 - u}{1 +u }\big)^p e^{-\frac{u^2}{2a^2}}\,du \\- &\int_{1}^{+\infty} \big(\frac{1-u}{1+u}\big)^p e^{-\frac{u^2}{2a^2}}\,du \bigg]\end{align*} } &     \parbox{0.25\textwidth}{\begin{align*}\mathbb{E}[Z^2] = 2\omega\bigg[&\int_{0}^{1} \big(\frac{1 - u}{1 +u  }\big)^{2p} e^{-\frac{u^2}{2a^2}}\,du \\&- \int_{1}^{+\infty} \big(\frac{1-u}{1+u}\big)^{2p} e^{-\frac{u^2}{2a^2}}\,du \bigg]\\\end{align*}} &\\\bottomrule[1.5pt]
                \end{tabular}%
    }
    \caption{Criteria expression, probability distribution and first two moments
    for IoU, GIoU, SIoU, and GSIoU. These are valid for the comparison of same
    size square boxes of width $\omega$ randomly shifted horizontally. Random
    shifts are sampled from a centered Gaussian distribution of variance
    $\sigma^2$ and $a=\sigma/\omega$.}
    \label{tab:pdf_crit}
    \end{table*}

\begin{proof}
    First, in the setup described in \cref{prop:distribution_giou}, GIoU can be
    expressed in terms of the width of the boxes $\omega$ and the shift in
    between $x$. Let's call this function $\mathcal{C}$: 
    \begin{align*}
        \mathcal{C} \colon \mathbb{R} &\to [-1,1] \\
                 x &\mapsto \frac{\omega- |x|}{\omega + |x|}
    \end{align*}

    \noindent
    The shifts are sampled from a Gaussian distribution: $X \sim \mathcal{N}(0,
    \sigma^2)$, therefore we are interested in the distribution of the variable
    $Z=\mathcal{C}(X)$.

    \noindent
    The cumulative density function of $Z$ is given by:

    \begin{align*}
        F_Z(z) = P(Z \leq z) &= P( \frac{\omega- |X|}{\omega + |X|} \leq y)\\
                            & = P( \omega\frac{1- z}{1+z} \leq |X|) \\
                            & = 2 P( \omega\frac{1- z}{1+z} \leq X) \\
                            & = 2 (1 - F_X(\omega\frac{1- z}{1+z})) \\
                            & = 2 (1 - F_X(g(z)))
    \end{align*}
    With $g(z) = \omega(\frac{1- z}{1+z})$.\\

    \noindent
    Hence, the density function of $Z$ can be derived by taking the derivative of $F_Z$:

    \begin{align*}
        d_Z(z) &= \frac{d}{dz} F_Z(z) \\
                &= -2 g'(z) F_X'(g(y)) \\
                &= \frac{4\omega}{(1+z)^2\sqrt{2\pi} \sigma} \exp\left(-\frac{1}{2}\left[\frac{\omega(1-z)}{\sigma(1+z)}\right]^2\right)
    \end{align*}

    \noindent
    To determine the first and second moments of $Z$, we make use of the change of
    variable formula: 
    \begin{align*}
        \mathbb{E}[Z] = \mathbb{E}[\mathcal{C}(X)] = \int_{-\infty}^{+\infty} \mathcal{C}(x)d_X(x)\,dx \\
        \mathbb{E}[Z^2] = \mathbb{E}[\mathcal{C}(X)^2] = \int_{-\infty}^{+\infty} \mathcal{C}(x)^2d_X(x)\,dx
    \end{align*}\\[2in]
    \noindent
    Let's start with $\mathbb{E}[Z]$:

    \begin{align*}
        \mathbb{E}[Z]& = \int_{-\infty}^{+\infty} \mathcal{C}(x)d_X(x)\,dx \\
                    &= \int_{-\infty}^{+\infty} \frac{1}{\sqrt{2\pi} \sigma}\frac{\omega- |x|}{\omega + |x| } e^{-\frac{x^2}{2\sigma^2}}\,dx \\
                    &= \frac{2}{\sqrt{2\pi} \sigma}\int_{0}^{+\infty}  \frac{\omega- x}{\omega + x}e^{-\frac{x^2}{2\sigma^2}}\,dx \\
                    &= \frac{2}{\sqrt{2\pi} \sigma}\int_{0}^{+\infty}  \frac{2\omega- (\omega + x)}{\omega + x}e^{-\frac{x^2}{2\sigma^2}}\,dx \\
                    &= \frac{2}{\sqrt{2\pi} \sigma}\left[2\int_{0}^{+\infty}  \frac{\omega}{\omega + x}e^{-\frac{x^2}{2\sigma^2}}\,dx - \int_{0}^{+\infty}  e^{-\frac{x^2}{2\sigma^2}}\,dx\right] \\
                    &= \frac{4}{\sqrt{2\pi} \sigma}\int_{0}^{+\infty}  \frac{\omega}{\omega + x}e^{-\frac{x^2}{2\sigma^2}}\,dx - 1 \\
                    &= \frac{4}{\sqrt{2\pi} a}\int_{0}^{+\infty}  \frac{1}{1 + u}e^{-\frac{u^2}{2a^2}}\,du - 1 \\
                    &= \frac{4}{\sqrt{2\pi} a} \frac{\sqrt{2}a}{2\pi} G^{2,3}_{3,2}\left(2a^2 \bigg\rvert \begin{matrix}0 & \frac{1}{2} & \frac{1}{2} \\ \frac{1}{2}& 0 \end{matrix}\right) - 1 \\
                    &= \frac{2}{\pi^{3/2}} G^{2,3}_{3,2}\left(2a^2 \bigg\rvert \begin{matrix}0 & \frac{1}{2} & \frac{1}{2} \\ \frac{1}{2}& 0 \end{matrix}\right) -1
    \end{align*}

    \noindent
    From line 2 to 3, we used the parity of function $\mathcal{C}$, between 6
    and 7, a change of variable $u=x/\omega$ is done and $a$ is set to $\sigma /
    \omega$. Finally, in the second-to-last line, we identify a Meijer-G
    function \cite{meijer2013} evaluated at $2a^2$.  Unfortunately, there exist
    no closed-form of the integral $ \int_{0}^{+\infty}  \frac{1}{1 +
    u}e^{-\frac{u^2}{2a^2}}\,dx$. In this case, a Mellin transform of this
    integral allows recognizing a Meijer-G function. For other criteria, their
    first two moments cannot be expressed in a similar closed form. That is why we only
    provide the theoretical expressions of the expectation and variance of GIoU.
    \\

    \noindent
    A similar derivation leads to the expression of the second moment of $Z$:
    \begin{align*}
        \mathbb{E}[Z^2]& = \int_{-\infty}^{+\infty} \mathcal{C}(x)^2d_X(x)\,dx \\
                    &= \int_{-\infty}^{+\infty} \frac{1}{\sqrt{2\pi} \sigma}\big(\frac{\omega- |x|}{\omega + |x| }\big)^2 e^{-\frac{x^2}{2\sigma^2}}\,dx \\
                    &= \frac{2}{\sqrt{2\pi} \sigma}\int_{0}^{+\infty}  \big(\frac{\omega- x}{\omega + x}\big)^2e^{-\frac{x^2}{2\sigma^2}}\,dx \\
                    &= 1 - \frac{8\omega}{\sqrt{2\pi} \sigma}\int_{0}^{+\infty}  \frac{x}{(\omega + x)^2}e^{-\frac{x^2}{2\sigma^2}}\,dx \\
                    &= 1 - \frac{8}{\sqrt{2\pi} a}\left[ a^2 -2\sigma^2\int_{0}^{+\infty}  \frac{1}{(\omega + x)^3}e^{-\frac{x^2}{2\sigma^2}}\,dx\right] \\
                    &= 1 - \frac{8}{\sqrt{2\pi} a}\left[ a^2 -2a^2\int_{0}^{+\infty}  \frac{1}{(1 + u)^3}e^{-\frac{u^2}{2a^2}}\,du\right] \\
                    &=1- \frac{8a}{\sqrt{2\pi}} + \frac{16a^2}{\pi^{3/2}} G^{2,3}_{3,2}\left(2a^2 \bigg\rvert \begin{matrix}-1 & \frac{1}{2} & -\frac{1}{2} \\ \frac{1}{2}  & 0 \end{matrix}\right)   
    \end{align*}

    \noindent
    From line 2 to 3, we again use the parity of $\mathcal{C}$, from 4 to 5, an
    integration by parts is done, and finally, from 5 to 6, we apply the change of
    variable $u=x/\omega$. Once again, we get an integral that does not have any
    closed form but can be expressed by another Meijer-G function. \\

    \noindent
    For completeness, we recall here the definition of the Meijer-G function: 

    \begin{align}
        &G^{m,n}_{p,q}\left(z \bigg\rvert \begin{matrix}a_1 & ... & a_p \\ b_1 & ... & b_q \end{matrix}\right)\\
        & =\frac{1}{2 \pi i} \int_{L}\frac{\prod\limits_{j=1}^m\Gamma(b_j - s)\prod\limits_{j=1}^n\Gamma(1 - a_j + s)}{\prod\limits_{j=m+1}^q\Gamma(1 - b_j + s)\prod\limits_{j=n+1}^p\Gamma(a_j - s)} z^s\, ds, \nonumber
    \end{align}

    where $L$ is the integration path and $\Gamma$ is the gamma function. $m$,
    $n$, $p$ and $q$ are integers while $a_j$ and $b_j$ are real or complex
    numbers. There are some constraints on these parameters, but we do not detail
    them here, they can be found in \cite{meijer2013}.

\end{proof}

\noindent
\subsection{Non-Closed-Form Expressions for moments of IoU, SIoU, and GSIoU} 
Other criteria do not have closed forms for their first and second moments.
Nonetheless, we provide in this appendix their expressions keeping the integrals
as simple as possible, which allows relatively easy evaluation. In addition, we
provide the expression of the pdf for each criterion. The setup remains
identical as in \cref{prop:distribution_giou}, the boxes, of the same width
$\omega$, are only horizontally shifted. For clarity, we also give simple
expressions of each criterion in such a setup (see \cref{tab:pdf_crit})

\section{User study}
\label{app:user_study}

\begin{figure*}
    \centering
    \includegraphics[width=\textwidth]{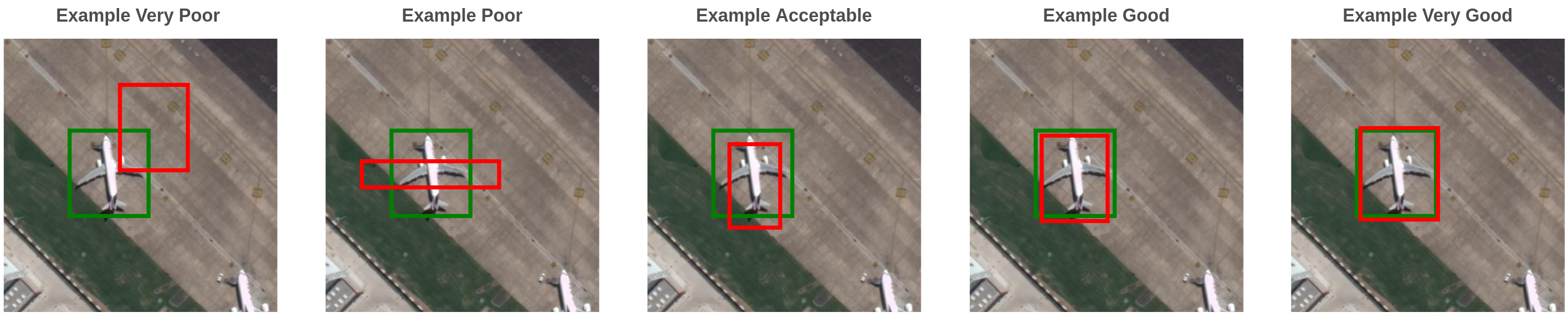}
    \caption{Examples given to participants of the user study. The IoU between
    the green and red boxes are 0.1, 0.25, 0.5, 0.75, and 0.9 for the ratings
    from very poor to very good respectively.}
    \label{fig:user_study_example}
\end{figure*}

\noindent
\textbf{Experimental Protocol} \\
To carry out the user study about detection preferences, we developed a Web App
to gather participants' answers. Each participant had to sign in with a brief
form. They are asked about their age and if they are familiar with image
analysis. After completing the form, participants are brought to the experiment
page. On this page, one image is visible with bounding boxes drawn on it. A
green one, which represents the ground truth annotation of an object, and a red
one randomly shifted and deformed. Each participant must rate how well the red
box is detecting the object inside the green box. The rating is done on a
5-levels scale, going from \textit{very poor} to \textit{very good}. A set of 50
different images is shown to each participant. After 25 images, the experiment
changes slightly: the background image is replaced by a completely black image.
This would remove any contextual bias coming from the variety of objects inside
the green box. We refer to the two phases of the experiment as the phases with
or without context respectively. The red boxes are sampled around the green box,
but to enforce a uniform distribution of the IoU with the green box, a random
IoU value $\psi $ is first uniformly sampled between 0 and 1. Then, we randomly
generate a red box that has an IoU $\psi$ with the green box (direct box
sampling does not produce uniformly distributed IoU values). Participants are
instructed to answer quickly and are provided with examples for each possible
rating (see \cref{fig:user_study_example}). The images and the annotations are
randomly picked from the dataset DOTA.\\

\noindent
\textbf{General Statistics about Participants} \\
The study gathered 75 participants and a total of 3136 individual answers (because some
participants did not complete the entire experiment). The age of the
participants ranges from 21 to 64 years old with an average of 31. Approximately
half (37) of the participants are knowledgeable in computer vision or image
analysis, we will refer to this group as the \textit{expert group}. On average,
the response time is 10.3s per evaluated image during the first phase of the experiment (when a
background image is visible). It drops to 7.2s when the image is replaced
by a uniform background during the second phase. This time difference suggests that humans do
take into account the contextual information of the image inside their
decision-making process. \\

\noindent
\textbf{User Study Insights} \\
We are interested here in several variables: the object size, the presence of
contextual information, the expertise, and the age of the participants. The
object size (small, medium, or large) is at the heart of our analysis, and we
want to study its influence on human rating. The other three variables are
studied mostly to explore possible biases in the experiment. The main result
coming from this study is that for the same IoU value, humans will attribute a
higher rating to smaller objects. This suggests that IoU is a too strict
criterion for small objects and that a better criterion should not be
scale-independent. This is reported in \cref{tab:avg_rating}, which gathers the
average rating $r$ under different groupings (by object size, presence of
contextual information, the expertise of the participants, and age of the
participants). In addition, the average value for each criterion is given for
each group. IoU value is close to 0.5 for every group as expected (boxes were
chosen to have a uniformly distributed IoU). However, values of other criteria
vary from one group to another. This is especially true for scale-dependent
criteria (SIoU and NWD) on different object size groups. To check whether the
different groups are statistically different, we conducted one-way ANOVA tests
on the four variables from \cref{tab:avg_rating}. The results confirm that the
mean ratings for various object sizes are statistically different ($p<8.4\times
10^{-26}$ \footnote{$p$ stands for the $p$-value of the statistical test here, not
the exponent from SIoU's definition.}). However, the tests find no statistical
difference for other variables: participant expertise ($p<0.47$), participant
age ($p<0.02$), and presence of contextual information ($p<0.28$). 

\begin{figure*}
    \centering
    \includegraphics[width=\textwidth]{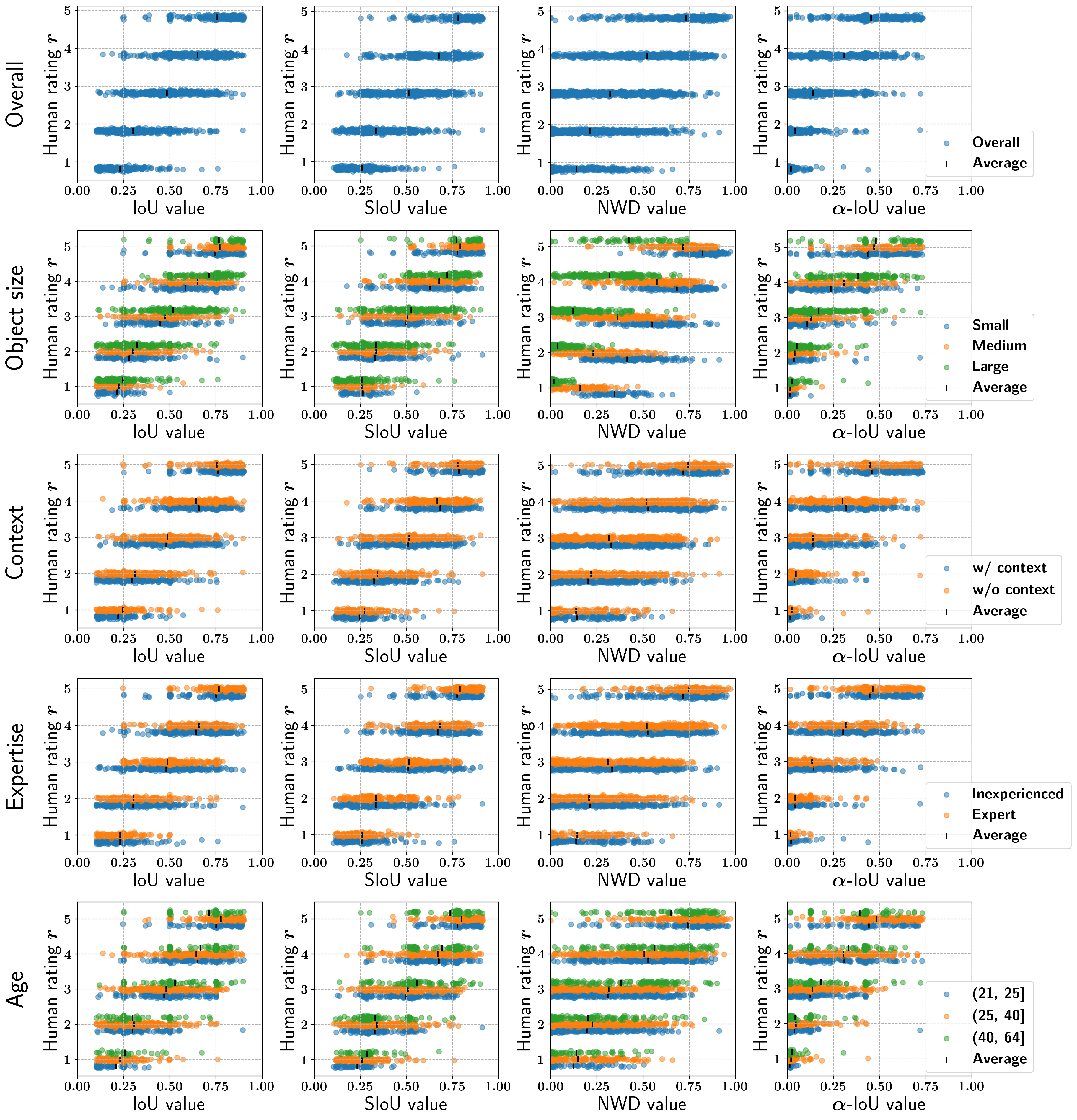}
    \caption{Rating against IoU, SIoU ($\gamma=0.2$, $\kappa=64$), NWD and
    $\alpha$-IoU ($\alpha=3$) values, overall and for different groupings of the
    variables of interest (object size, presence of contextual information,
    expertise and age of the participants). Colors represent different values for
    each variable, a legend for each row is included in the right-most column
    of the figure. For the \textit{Age} variable, the participants have been
    separated into three groups of the same size. }
    \label{fig:hr_vs_crit}
    \vspace{2cm}
\end{figure*}

\begin{figure*}
    \centering
    \includegraphics[width=0.95\textwidth]{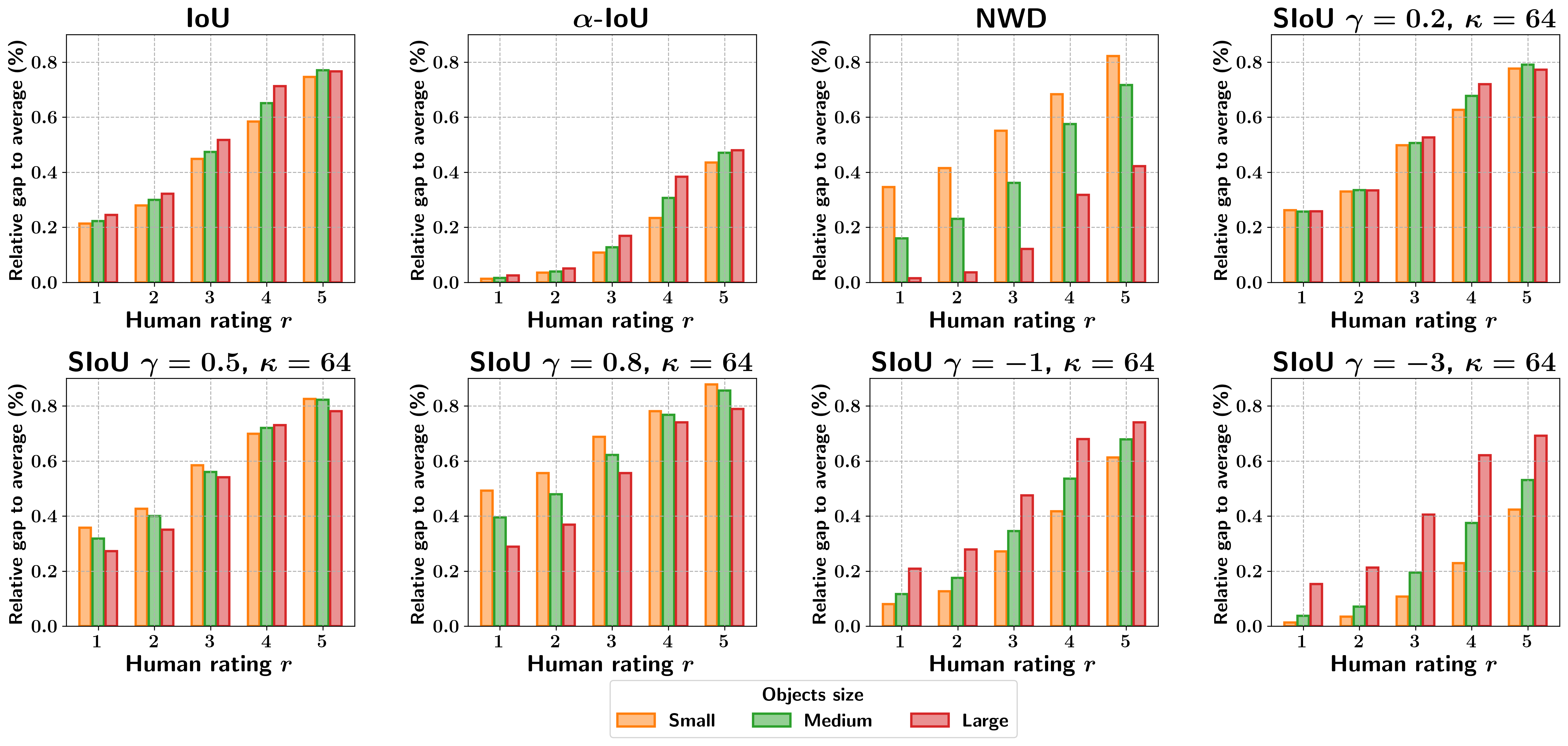}
    \caption{Criteria' scores for different object sizes and human ratings $r
    \in \{1,2,3,4,5\}$. Multiple values of $\gamma$ are employed for SIoU to
    highlight its influence on the criterion values.}
    \label{fig:bar_chart_user_study}
\end{figure*}

To investigate the alignment of the various criteria with the human perception,
\cref{fig:hr_vs_crit} plots the rating values against the criteria value. For
visualization purposes, random vertical shifts were added to rating values to
distinguish between the values of each variable and between data points. From
this figure, it is clear that the IoU is not a perfect criterion as a wide range
of IoU values is attributed to the same rating value. It seems also clear that
contextual information, participant expertise, and age do not introduce much
change in the human rating. However, the object size has a consistent influence
on the rating: the average IoU value for a rating value decreases with the
object size (these are visible as the black vertical lines in
\cref{fig:rating_vs_crit}). This completely agrees with the statistical test
results. SIoU can compensate for this trend and produces more aligned
averages for the different object sizes. NWD has the same effect, but largely
reverses the trend in the other direction. Finally, the exact values of the
averages for each object size and rating value are available in
\cref{tab:avg_iou_object_size}. These are also presented in the form of bar
charts in \cref{fig:rating_vs_crit,fig:bar_chart_user_study}.
\cref{fig:bar_chart_user_study} presents similar bar charts for all criteria and
even for SIoU with various values of $\gamma$.

Finally, we also investigate the correlation between the various criteria and
human rating. The results of this analysis are available in
\cref{tab:correlation_full}. We choose Kendall's $\tau$ to deal with human
rating correctly as it is an ordered categorical variable. SIoU correlates best
with human rating. Of course, the correlation between the criteria is high as
they are mostly derivated from IoU, except NWD which shows a lower correlation
both with other criteria and human rating.

\begin{table}[h]
    \centering
    \resizebox{\columnwidth}{!}{%
    \begin{tabular}{@{}ccccccc@{}}
    \toprule[1.5pt]
    \textbf{}                             & \textbf{}     & $\boldsymbol{r}$ & \textbf{IoU} & \textbf{SIoU} & \textbf{NWD} & $\boldsymbol \alpha$\textbf{-IoU} \\ \midrule[1.5pt]
    \multirow{3}{*}{\textbf{Object size}} & Small         & 3.406                            & 0.507                             & 0.550                            & 0.610                                   & 0.203                \\
                                          & Medium        & 3.158                            & 0.502                             & 0.532                            & 0.424                                   & 0.199                \\
                                          & Large         & 2.824                            & 0.491                             & 0.500                            & 0.151                                   & 0.189                \\\midrule
    \multirow{2}{*}{\textbf{Context}}     & w/o context   & 3.144                            & 0.504                             & 0.531                            & 0.397                                   & 0.197                \\
                                          & w/ context    & 3.112                            & 0.496                             & 0.523                            & 0.390                                   & 0.197                \\\midrule
    \multirow{2}{*}{\textbf{Expertise}}   & Inexperienced & 3.104                            & 0.493                             & 0.520                            & 0.392                                   & 0.194                \\
                                          & Expert        & 3.152                            & 0.507                             & 0.535                            & 0.395                                   & 0.200                \\\midrule
    \multirow{3}{*}{\textbf{Age}}         & (10, 25]      & 3.215                            & 0.504                             & 0.531                            & 0.397                                   & 0.196                \\
                                          & (25, 40]      & 3.078                            & 0.496                             & 0.524                            & 0.390                                   & 0.198                \\
                                          & (40, 65]      & 3.085                            & 0.501                             & 0.529                            & 0.394                                   & 0.197                \\ \bottomrule[1.5pt]
    \end{tabular}%
    }
    \caption{Average rating and criteria values for different groupings of the
    variables of interest (object size, presence of contextual information,
    expertise and age of the participants).}
    \label{tab:avg_rating}
    \end{table}

\begin{table}[h]
    \centering
    \resizebox{0.85\columnwidth}{!}{%
    \begin{tabular}{@{}cccccc@{}}
    \toprule[1.5pt]
    \textbf{}                        & $\boldsymbol{r}$ & \textbf{IoU} & \textbf{SIoU} & \textbf{NWD} & $\boldsymbol \alpha$\textbf{-IoU} \\ \midrule[1.5pt]
    \multirow{5}{*}{\parbox{2cm}{\textbf{Small \\Objects}}}  & 1         & 0.214        & 0.262         & 0.346        & 0.013               \\
                                     & 2         & 0.279        & 0.330         & 0.415        & 0.035               \\
                                     & 3         & 0.449        & 0.498         & 0.551        & 0.108               \\
                                     & 4         & 0.584        & 0.627         & 0.683        & 0.234               \\
                                     & 5         & 0.746        & 0.776         & 0.822        & 0.435               \\ \midrule
    \multirow{5}{*}{\parbox{2cm}{\textbf{Medium \\Objects}}} & 1         & 0.223        & 0.257         & 0.160        & 0.016               \\
                                     & 2         & 0.299        & 0.335         & 0.230        & 0.039               \\
                                     & 3         & 0.474        & 0.506         & 0.361        & 0.127               \\
                                     & 4         & 0.651        & 0.677         & 0.575        & 0.306               \\
                                     & 5         & 0.771        & 0.791         & 0.716        & 0.470               \\\midrule
    \multirow{5}{*}{\parbox{2cm}{\textbf{Large \\Objects}}}  & 1         & 0.245        & 0.258         & 0.015        & 0.025               \\
                                     & 2         & 0.322        & 0.334         & 0.036        & 0.051               \\
                                     & 3         & 0.517        & 0.527         & 0.121        & 0.168               \\
                                     & 4         & 0.713        & 0.720         & 0.318        & 0.383               \\
                                     & 5         & 0.766        & 0.772         & 0.422        & 0.480               \\\bottomrule[1.5pt]
    \end{tabular}%
    }
    \caption{Average criteria (IoU, SIoU, NWD and $\alpha$-IoU) values for different object sizes and human rating values.}
    \label{tab:avg_iou_object_size}
    \end{table}

\begin{table}[h]
    \vspace{-2mm}
    \centering
    \resizebox{0.7\columnwidth}{!}{%
    \begin{tabular}{@{}cccccc@{}}
    \toprule
    \textbf{}                   & $\boldsymbol{r}$& \textbf{IoU} & \textbf{SIoU} & $\boldsymbol{\alpha}$\textbf{-IoU} & \textbf{NWD} \\ \midrule
    $\boldsymbol{r}$          & 1.000           & 0.674        & \textbf{0.701}& 0.674               & 0.550        \\ \bottomrule
    \textbf{IoU}                & 0.674           & 1.000        & 0.892         & 0.997               & 0.474        \\
    \textbf{SIoU}               & 0.701           & 0.892        & 1.000         & 0.892               & 0.576        \\
    $\boldsymbol{\alpha}$\textbf{-IoU}   & 0.674           & 0.997        & 0.892         & 1.000               & 0.472        \\
    \textbf{NWD}                & 0.550           & 0.474        & 0.576         & 0.472               & 1.000        \\ \bottomrule
    \end{tabular}%
    }
    \caption{Kendall's $\tau$ correlation between various criteria and human
    rating $r$. For SIoU, $\gamma=0.2$ and $\kappa=64$, for $\alpha$-IoU,
    $\alpha=3$.}
    \label{tab:correlation_full}
\end{table}

\section{Few-Shot Experiments}
\label{app:experimental_results}
In this appendix, we provide the implementation details for our FSOD
experiments. Some additional results with different FSOD methods are also included in
\cref{tab:fsod_method_com}.

\subsection{Implementation details}
\label{app:implem_details}
Our few-shot experiments are based on Cross-scale Query Support Alignment (XQSA)
\cite{lejeune2022aaf}, a recently proposed method for few-shot object detection.
Except for experiments from \cref{tab:fsod_method_com}, all the few-shot
experiments were done with XQSA. This method is based on an attention mechanism
and combines features from support and query images to condition the detection
of the selected classes. This method leverages a ResNet-50 as a backbone with a 3-layers Feature
Pyramid Network on top. The detection head is based on FCOS and detects
independently objects from different classes. The training of all FSOD methods
employed in this paper is done episodically. For each episode, a subset
$\mathcal{C}_{ep} \subset \mathcal{C}$ of the classes is randomly sampled
($\lvert\mathcal{C}_{ep} \rvert= 3$). Only the annotations of the episode
classes are used for training the model. A support set is sampled at each
episode containing $K$ examples for each episode class. A query set is also
sampled for each episode, containing 100 images per class. The query set is used
to compute the losses and update model's weights. 

The training is divided into \textit{base training} and \textit{fine-tuning}.
During base training, only base classes can be sampled ($\mathcal{C}_{ep}
\subset \mathcal{C}_{base}$) and one example per class is drawn for the support
set ($K=1$). The optimization is done with SGD and a learning rate of
$\num{1e-3}$ for 1000 episodes. During \textit{fine-tuning}, the backbone is
frozen, the learning rate is divided by 10 and the episode classes can be
sampled from $\mathcal{C}_{base} \cup \mathcal{C}_{novel}$, with two novel
classes per episode. Examples from novel classes are selected among the $K=10$
examples sampled once before fine-tuning. The base/novel splits are selected
randomly before training. 3 classes are selected as novel classes for DOTA, 5
for DIOR and Pascal VOC, and 20 for COCO. Note that for COCO, the split is not
chosen randomly, but instead the 20 novel classes are the ones that overlap with
Pascal VOC. 

The evaluation is also conducted in an episodic manner. The performance is
averaged over multiple episodes each containing 500 examples for each class and
this is repeated 10 times with randomly sampled support sets. The query
and support examples are drawn from test set, thus the support examples are
different from the ones used during fine-tuning. This prevents overestimations
of the performance due to overfitting on the support examples. 

\subsection{Few-Shot Methods Comparison}
\label{app:fs_comparison}
To support the versatility of GSIoU, we also experiment with several few-shot
approaches. We select three existing FSOD techniques Feature
Reweighting \cite{kang2019few} (FRW),  Dual-Awarness Atention
\cite{chen2021should} (DANA) and Cross Scale Query Support Alignment \cite{lejeune2022aaf}
(XQSA). We compare GIoU and GSIoU as regression losses.
\cref{tab:fsod_method_com} regroups the results and shows consistent
improvements with GSIoU over GIoU. Except for DANA, GSIoU improves detection
performance, especially for small objects. 

\begin{table}[ht]
    \centering
    \resizebox{\columnwidth}{!}{%
    \begin{tabular}{@{}ccccccccccc@{}}
    \toprule[1.5pt]
    \textbf{}                      & \textbf{} & \multicolumn{4}{c}{\textbf{Base classes}}          & & \multicolumn{4}{c}{\textbf{Novel Classes}}          \\ 
    \textbf{}                      & \textbf{XQSA} & \textbf{All} & \textbf{S} & \textbf{M} & \textbf{L}& & \textbf{All} & \textbf{S} & \textbf{M} & \textbf{L} \\ \midrule[1.5pt]
    \multirow{2}{*}{\textbf{FRW}}  & w/ GIoU      & \textbf{34.60}        & \textbf{16.15}      & \textbf{48.61}      & \textbf{59.00}     & & 32.00        & 15.29      & \textbf{44.50}      & 54.77      \\
                                   & w/ GSIoU     & 30.36        & 11.94      & 44.30      & 54.87     & & \textbf{32.94}        & \textbf{16.69}      & 42.87      & \textbf{62.64}      \\ \midrule
    \multirow{2}{*}{\textbf{DANA}} & w/ GIoU      & 48.09        & 27.34      & \textbf{66.06}      & \textbf{68.00}     & & \textbf{44.49}        & \textbf{30.10}      & 52.24      & 74.40      \\
                                   & w/ GSIoU     & \textbf{50.10}        & \textbf{32.19}      & 65.46      & 67.77     & & 41.40        & 21.07      & \textbf{54.80}      & \textbf{75.23}      \\ \midrule
    \multirow{2}{*}{\textbf{XQSA}} & w/ GIoU      & \textbf{45.30}        & \textbf{26.94}      & 61.17      & 63.00     & & 41.03        & 24.01      & \textbf{52.13}      & 69.78      \\
                                   & w/ GSIoU     & 43.42        & 22.14      & \textbf{61.19}      & \textbf{66.02}     & & \textbf{45.88}        & \textbf{34.83}      & 51.26      & \textbf{70.78}      \\ \bottomrule[1.5pt]
    \end{tabular}%
    }
    \caption{Performance comparison with three different FSOD methods: Feature
    Reweighting \cite{kang2019few} (FRW), Dual Awarness Attention
    \cite{chen2021should} (DANA) and Cross-scale Query Support Alignment
    \cite{lejeune2022aaf} (XQSA), trained with GIoU and GSIoU. mAP is reported
    with a 0.5 \textbf{IoU threshold} for small (S), medium (M), large (L) and
    all objects.}
    \label{tab:fsod_method_com}
    \end{table}

\section{Evaluation with SIoU}
\label{app:siou_eval}
In this appendix, we present the results from \cref{sec:results} but using SIoU as
the evaluation criterion. Specifically, instead of choosing an IoU threshold to
decide if a box is a positive or negative detection, an SIoU threshold is
employed. For the sake of comparison, we kept the same thresholds as in
\cref{tab:criterion_comparison,tab:dataset_comparison,tab:regular_od}, \ie 0.5
for Few-Shot methods and 0.5:0.95 for regular object detection. The results are
available in
\cref{tab:criteria_comp_siou,tab:dataset_comparison_siou,tab:regular_od_siou}.
The conclusions from \cref{sec:results} still hold, and the superiority of GSIoU
over other criteria is clear. However, a few changes are noticeable. First, SIoU loss
seems to perform better than IoU. This is expected, the model is directly
optimized to satisfy this criterion.
Then, when evaluated with SIoU, models trained with NWD perform well.
Indeed, NWD puts a lot of emphasis on size matching during
training, and less on position. Therefore, it is logical to observe better
performance compared to other losses when using SIoU as the evaluation criterion. 

One crucial point is that SIoU evaluation mostly changes the score for small
objects. SIoU behaves like IoU for large objects, therefore relatively small
changes are visible for medium and large objects. Overall, the scores are higher
than with IoU as the expected value of SIoU is higher than IoU. More boxes are
considered positive detection. But what is important to note is that the gap
between small and large objects performance is reduced and therefore aligns
better with human perception.

\makeatletter
\setlength{\@fptop}{0pt}
\setlength{\@fpsep}{20pt}
\makeatother

\begin{table}[ht]
    \centering
    \resizebox{\columnwidth}{!}{%
    \begin{tabular}{@{}cccccccccc@{}}
    \toprule[1.5pt]
                       & \multicolumn{4}{c}{\textbf{Base classes}}          & & \multicolumn{4}{c}{\textbf{Novel Classes}}                           \\
    \textbf{Loss}      & \textbf{All} & \textbf{S} & \textbf{M} & \textbf{L}& & \textbf{All} & \textbf{S} & \textbf{M} & \textbf{L}  \\  \midrule[1.5pt]
    \textbf{IoU}       & 55.81        & 35.03      & 62.57      & 70.05     & & 39.10        & 18.58      & 53.93      & 68.83           \\
    \textbf{$\boldsymbol \alpha$-IoU} & 53.05        & 20.60      & 61.05     & \textbf{72.41}     & & 41.93        & 20.99      & 55.74   & 76.79         \\
    \textbf{SIoU}      & \textbf{59.77}        & 36.38      & 67.29      & 70.06     & & 49.51        & 31.06      & 62.53      & \textbf{77.24}           \\ \midrule
    \textbf{NWD}       & 58.80        & 34.16      & 66.81      & 70.05     & & 53.66        & 42.02      & 62.53      & 68.92           \\
    \textbf{GIoU}      & 59.27        & \textbf{44.07}      & \textbf{66.91}      & 65.46     & & 49.02        & 35.10      & 57.58      & 74.30           \\
    \textbf{GSIoU}     & 59.32        & 35.32      & 66.29      & 69.03     & & \textbf{57.70}        & \textbf{46.77}      & \textbf{65.56}      & 73.67           \\ \bottomrule[1.5pt]
    \end{tabular}%
    }
    \caption{Few-shot performance comparison between several criteria: IoU,
    $\alpha$-IoU, SIoU, NWD, GIoU and GSIoU trained on DOTA. mAP is reported
    with a 0.5 \textbf{SIoU threshold} for small (S), medium (M), large (L), and all objects.}
    \label{tab:criteria_comp_siou}
    \end{table}

\begin{table}[t!]
        \centering
        \resizebox{0.9\columnwidth}{!}{%
        \begin{tabular}{@{}cccccccccc@{}}
        \toprule[1.5pt]
                                           & \multicolumn{4}{c}{\textbf{DOTA}}                                                                                           & \multicolumn{4}{c}{\textbf{DIOR}}                                                                                           \\ \midrule
        \textbf{FCOS}                          & \textbf{All}             & \multicolumn{1}{c}{\textbf{S}} & \multicolumn{1}{c}{\textbf{M}} & \multicolumn{1}{c}{\textbf{L}}& & \textbf{All}             & \multicolumn{1}{c}{\textbf{S}} & \multicolumn{1}{c}{\textbf{M}} & \multicolumn{1}{c}{\textbf{L}} \\
        \multicolumn{1}{c}{\textbf{w/ GIoU}}  & \multicolumn{1}{r}{43.9} & 27.4                           & 46.5                           & 47.2                          & & \multicolumn{1}{r}{54.5} & 17.6                           & 49.8                           & 66.4                           \\
        \multicolumn{1}{c}{\textbf{w/ GSIoU}} & \multicolumn{1}{r}{\textbf{45.4}} & \textbf{27.7}                           & \textbf{50.2}                           & \textbf{49.2}                          & & \multicolumn{1}{r}{\textbf{55.4}} & \textbf{18.0}                             & \textbf{50.1}                           & \textbf{69.2}                           \\ \bottomrule[1.5pt]
        \end{tabular}%
        }
        \caption{Regular Object Detection performance on DOTA and DIOR datasets with
        GIoU and GSIoU ($\gamma=-3$ and $\kappa=16$) losses. mAP is computed with
        several \textbf{SIoU thresholds} (0.5 to 0.95) as it is commonly done in regular
        detection.}
        \label{tab:regular_od_siou}
        \end{table}

\begin{table}[t!]
    \centering
    \resizebox{\columnwidth}{!}{%
    \begin{tabular}{@{}ccccccccccc@{}}
    \toprule[1.5pt]
    \textbf{}                        & \textbf{} & \multicolumn{4}{c}{\textbf{Base classes}}          & & \multicolumn{4}{c}{\textbf{Novel Classes}} \\ 
    \textbf{}                        & \textbf{XQSA} & \textbf{All} & \textbf{S} & \textbf{M} & \textbf{L}& & \textbf{All}  & \textbf{S}  & M     & L    \\\midrule[1.5pt]
    \multirow{2}{*}{\textbf{DOTA}}   &  w/ GIoU      & 59.27        & \textbf{44.07}      & 66.91      & 65.46     & & 49.02        & 35.10      & 57.58      & \textbf{74.30}      \\
                                     &  w/ GSIoU     & \textbf{59.32} & 35.32      & \textbf{66.29}      & \textbf{69.03}     & & \textbf{57.70}        & \textbf{46.77}      & \textbf{65.56}      & 73.67      \\\midrule
    \multirow{2}{*}{\textbf{DIOR}}   &  w/ GIoU      & 62.06        & 17.49      & 45.55      & 82.22     & & 53.81        & 23.79      & 53.46      & 71.63      \\
                                     &  w/ GSIoU     & \textbf{63.81}        & \textbf{17.77}      & \textbf{49.62}      & \textbf{82.53}     & & \textbf{58.79}        & \textbf{25.60}      & \textbf{59.28}      & \textbf{73.78}      \\\midrule
    \multirow{2}{*}{\textbf{Pascal}} &  w/ GIoU      & 55.51        & 26.10      & \textbf{46.82}      & 64.31     & & 52.43        & 28.97      & 40.73      & 62.58      \\
                                     &  w/ GSIoU     & \textbf{58.74}        & \textbf{27.47}      & 46.56      & \textbf{68.93}     & & \textbf{58.92}        & \textbf{31.36}      & \textbf{41.65}      & \textbf{69.71}      \\ \midrule
    \multirow{2}{*}{\textbf{COCO}} &  w/ GIoU      & 21.46                 & 12.77               & 24.79               & 31.86              & & 29.21          & 17.36         & 27.62         & 40.05   \\
                                   &  w/ GSIoU     & \textbf{21.97}        & \textbf{12.80}      & \textbf{25.72}      & \textbf{32.35}     & & \textbf{29.94} & \textbf{18.87}& \textbf{29.93}& \textbf{40.47}   \\ \bottomrule[1.5pt]
    \end{tabular}%
    }
    \caption{Few-shot performance on three datasets: DOTA, DIOR, Pascal VOC and COCO.
    GIoU and GSIoU losses are compared. mAP is reported with a 0.5 \textbf{SIoU threshold}
    and for various object sizes.}
    \label{tab:dataset_comparison_siou}
    \end{table}